\pdfoutput=1

\documentclass{article}

\usepackage{arxiv}

\usepackage{graphicx}
\usepackage{comment}
\usepackage{amsmath,amssymb}
\usepackage{color}

\usepackage{amsmath,amsfonts,bm}

\newcommand\widebar[1]{\mathop{\overline{#1}}}

\def\eqref#1{equation~\ref{#1}}

\def\1{\bm{1}}

\DeclareMathAlphabet{\mathsfit}{\encodingdefault}{\sfdefault}{m}{sl}
\SetMathAlphabet{\mathsfit}{bold}{\encodingdefault}{\sfdefault}{bx}{n}

\newcommand{\E}{\mathbb{E}}
\newcommand{\Ls}{\mathcal{L}}

\DeclareMathOperator*{\argmin}{arg\,min}

\usepackage{floatrow}
\usepackage{subfig}
\usepackage{graphicx}
\usepackage{hyperref}

\usepackage[title]{appendix}

\title{Meta-Learning Initializations for Image Segmentation}

\author{ Sean M.~Hendryx \\
  School of Information\\
  University of Arizona\\
  Tucson, AZ 85721, USA\\
  \texttt{seanmhendryx@email.arizona.edu}\\
  \And
  Andrew B. Leach \thanks{Now at Google.}\\
  Program in Applied Mathematics\\
  University of Arizona\\
  Tucson, AZ 85721, USA\\
  \texttt{ableach@email.arizona.edu}\\
  \AND
  Paul D. Hein \\
  School of Information\\
  University of Arizona\\
  Tucson, AZ 85721, USA\\
  \texttt{pauldhein@email.arizona.edu}\\
  \And
  Clayton T. Morrison \\
  School of Information\\
  University of Arizona\\
  Tucson, AZ 85721, USA\\
  \texttt{claytonm@email.arizona.edu}\\
}

\renewcommand{\headeright}{}
\renewcommand{\undertitle}{}

\begin{document}
\maketitle

\begin{abstract}
We extend first-order model agnostic meta-learning algorithms (including FOMAML and Reptile) to image segmentation, present a novel neural network architecture built for fast learning which we call EfficientLab, and leverage a formal definition of the test error of meta-learning algorithms to decrease error on out of distribution tasks. We show state of the art results on the FSS-1000 dataset by meta-training EfficientLab with FOMAML and using Bayesian optimization to infer the optimal test-time adaptation routine hyperparameters. We also construct a small benchmark dataset, FP-k, for the empirical study of how meta-learning systems perform in both few- and many-shot settings.
On the FP-k dataset, we show that meta-learned initializations provide value for canonical few-shot image segmentation but their performance is quickly matched by conventional transfer learning with performance being equal beyond 10 labeled examples.
Our code, meta-learned model, and the FP-k dataset are available at \url{https://github.com/ml4ai/mliis}.
\end{abstract}

\section{Introduction}
\label{Introduction}
In recent years, there has been substantial progress in high accuracy image segmentation in the high data regime (see \cite{liu2019auto} and their references).  
While meta-learning approaches that utilize neural network representations have made progress in few-shot image classification, reinforcement learning, and, more recently, image semantic segmentation, the training algorithms and model architectures have become increasingly specialized to the low data regime. A desirable property of a learning system is one that effectively applies knowledge gained from a few \textit{or} many examples, while reducing the generalization gap when trained on little data and not being encumbered by its own learning routines when there are many examples. This property is desirable because training and maintaining multiple models is more cumbersome than training and maintaining one model. A natural question that arises is how to develop learning systems that scale from few-shot to many-shot settings while yielding competitive accuracy in both. One scalable potential approach that does not require ensembling many models nor the computational costs of relation networks, is to meta-learn an initialization such as via Model Agnostic Meta-Learning (MAML)~\cite{finn2017model}.

In this work, we specifically address the problem of meta-learning initializations for deep neural networks that must produce dense, structured output, such as for the semantic segmentation of images. We ask the following questions:
\begin{enumerate}
  \item Do first-order MAML-type algorithms extend to the higher dimensional parameter spaces, dense prediction, and skewed distributions required of semantic segmentation?
  \item How sensitive is the test-time performance of gradient-based meta-learning to the hyperparameters of the update routine used to adapt the initialization to new tasks?
  \item What is the number of labeled examples beyond which a conventional approach to training deep neural networks matches or outperforms the meta-learned initializations?
\end{enumerate}

To address the third question, we put together a small benchmark dataset, which we call FP-k, that contains 5 tasks each with at least 420 labeled examples of image-mask pairs. In recent works~\cite{li2017meta,shaban2017one,rusu2018meta,zhang2019canet,lee2019meta}, few-shot learning approaches have become increasingly complex and specialized to the few-shot domain. Such specialization comes at an engineering cost and leaves many open questions, such as: What is the accuracy of a few-shot learning system when more labeled examples become available? After a certain number of labeled examples, will the few-shot learning system have the same accuracy as a simpler training approach such as conventional training via SGD? If so, what is the number of labeled examples beyond which a conventional approach to training deep neural networks matches or outperforms a meta-learning system? We address these questions in \ref{results}.

In summary, we address the above research questions as follows: We show that MAML-type algorithms do extend to few shot image segmentation, yielding state of the art results when their update routine is optimized after meta-training and when the model is regularized. Addressing question 2, we find that the meta-learned initialization's performance when being evaluated on a task is particularly sensitive to changes in the update routine's hyperparameters (see Figure \ref{fig:iou-over-lr}). We show theoretically in section \ref{motivating_UHO} and empirically in our results (see Table \ref{table:fss}) that a single update routine used both during meta-training and meta-testing may not have optimal generalization. Finally, we address question 3 by showing that our meta-learned initializations outperform ImageNet~\cite{deng2009imagenet} and joint-trained initializations when the number of labeled examples is less than 10. Our code, meta-learned model, and the FP-k dataset are available at \url{https://github.com/ml4ai/mliis}.

\section{Related Work}
\label{related}
Learning useful models from a small number of labeled examples of a new concept has been studied for decades~\cite{thrun1996learning} yet remains a challenging problem with no semblance of a unified solution. The advent of larger labeled datasets containing examples from many distinct concepts~\cite{vinyals2016matching} has enabled progress in the field in particular by enabling approaches that leverage the representations of nonlinear neural networks. Image segmentation is a well-suited domain for advances in few-shot learning given that the labels are particularly costly to generate~\cite{wei2019fss}.

Recent work in few-shot learning for image segmentation has utilized three key components: (1) model ensembling~\cite{shaban2017one}, (2) the relation networks of \cite{sung2018learning}\footnote{Not to be confused with the relation networks of \cite{santoro2017simple}.}, and (3) late fusion of representations~\cite{rakelly2018conditional,zhang2019canet,wei2019fss}. The inference procedure of ensembling models with a separately trained model for each example has been shown to produce better predictions than single shot approaches but will scale linearly in time and/or space complexity (depending on the implementation) in the number of training examples, as implemented in \cite{shaban2017one}.
 The use of multiple passes through subnetworks via iterative optimization modules was shown by  \cite{zhang2019canet} to yield improved segmentation results but comes at the expense of additional time complexity during inference.
The relation networks proposed in \cite{sung2018learning} have seen increased adoption in meta-learning systems and were recently extended to the modality of dense prediction by the authors in \cite{zhang2019canet} and \cite{wei2019fss}.
While this extension of the relation networks of \cite{sung2018learning} to image segmentation yield impressive results in the few-shot domain, their efficacy in scaling as more training data becomes available is untested.

Model Agnostic Meta-Learning (MAML) is a gradient-based meta-learning approach introduced in \cite{finn2017model}. First Order MAML (FOMAML) reduces the computational cost by not requiring backpropogating the meta-gradient through the inner-loop gradient and has been shown to work similarly well on classification tasks ~\cite{finn2017model, nichol2018reptile}. Though learning an initialization has the potential to unify few-shot and many-shot domains, initializations learned from MAML-type algorithms have been seen to overfit in the low-shot domain when adapting sufficiently expressive models such as deep residual networks that may be more than a small number of convolutional layers \footnote{The original MAML and Reptile convolutional neural networks (CNNs) use four convolutional layers with 32 filters each for MiniImagenet~\cite{finn2017model,nichol2018reptile}} ~\cite{mishra2018a,rusu2018meta}. The Meta-SGD learning framework added additional capacity to the same network architecture used in MAML with improved generalization by meta-learning a learning rate for each parameter in the network ~\cite{li2017meta}, but lacks a first order approximation. In addition to possessing potential to unify few- and many-shot domains, MAML-type algorithms are intriguing in that they impose no constraints on model architecture, given that the output of the meta-learning process is simply an initialization. Futhermore, the meta-learning dynamics, which learn a temporary memory of a sampled task, are related to the older idea of fast weights~\cite{hinton1987using,ba2016using}. Despite being dataset size and model architecture agnostic, MAML-type algorithms are unproven for high dimensionality of the hypothesis spaces and the skewed distributions of image segmentation problems data~\cite{rakelly2018conditional}.

In recent work, \cite{dhillon2019baseline} found that standard joint pre-training on all meta-training tasks on mini-imagenet, tiered ImageNet, and other few shot image \textit{classification} benchmarks, with a sufficiently large network is on par with many sophisticated few-shot learning algorithms. Furthermore, implementing meta-training code comes with additional complexity. Thus it is worth testing how a vanilla training loop on the ``joint'' distribution of all the 760 non-test tasks compares to a meta-learned initialization.

\section{Preliminaries}
\label{prelim}

\subsection{Generalization Error in Meta-learning}
In the context of image segmentation, an example from a task $\tau$ is comprised of an image $x$ and its corresponding binary mask $y$, which assigns each pixel membership to the target (ex. black bear) or background class. Examples $(x, y)$ from the domain $\mathcal{D}_{\tau}$ are distributed according to $q_{\tau}(x, y)$, and we measure the loss $\Ls$ of predictions $\hat{y}$ generated from parameters $\theta$ and a learning algorithm $U$. For a distribution $p(\tau)$ over the domain of tasks $\mathcal{T}$, the parameters that minimize the expected loss are

\begin{equation}
    \label{eq:1}
    \theta^{*} = \argmin_{\theta} \E_{p} \left[ \E_{q_{\tau}} \left[ \Ls \left( U (\theta)  \right) \right] \right]
\end{equation}

In practice, we only have access to a finite subset of the tasks, which we divide into the training $\mathcal{T}^{tr}$, validation $\mathcal{T}^{val}$, and test tasks $\mathcal{T}^{test}$, and instead optimize over an empirical distribution $\hat{p}(\tau) := p(\tau | \tau \in \mathcal{T}^{tr} )$.
For examples within each available task, we can similarly define $\mathcal{D}_{\tau}^{tr}$, $\mathcal{D}_{\tau}^{val}$, $\mathcal{D}_{\tau}^{test}$, and the corresponding empirical distribution $\hat{q}_{\tau}(x, y) := q_{\tau}(x, y | (x, y) \in \mathcal{D}_{\tau}^{tr} )$.
As a corollary to \ref{eq:1}, the empirically optimal initialization

\begin{equation}
	\label{eq:empirical_objective}
	\hat{\theta}^{*} = \argmin_{\theta} \E_{\hat{p}} \left[ \E_{\hat{q}_{\tau}} \left[ \Ls \left( U (\theta)  \right) \right] \right]
\end{equation}

has a generalization gap that can then be expressed as

\begin{equation}
	\E_{p} \left[ \E_{q_{\tau}} \left[ \Ls \left( U (\hat{\theta}^{*})  \right) \right] \right] - \E_{\hat{p}} \left[ \E_{\hat{q}_{\tau}} \left[ \Ls \left( U (\hat{\theta}^{*})  \right) \right] \right]
\end{equation}

We include a proof in the supplementary material. The generalization gap between the actual and empirical error in meta-learning is twofold: from the domain of all tasks $\mathcal{T}$ to the sample $\mathcal{T}^{tr}$, and within that, from all examples in $\mathcal{D}_{\tau}$ to $\mathcal{D}_{\tau}^{tr}$.

\subsection{Model Agnostic Meta-learning}
The model agnostic meta-learning (MAML) algorithm introduced in \cite{finn2017model} uses a gradient-based update procedure $U$ with hyperparameters $\omega$, which applies a limited number of training steps with a few-shot subset of $\mathcal{D}_{\tau}^{tr}$ to adapt a meta-learned initialization $\theta$ to each task. To be precise, $U$ maps from an initialization $\theta$ and examples in $\mathcal{D}_{\tau}$ to updated parameters $\theta_{\tau}$ which parameterize a task-specific prediction function $f(x;~\theta_{\tau})$:
\begin{equation}
  \label{eq:f}
  f(x;~\theta_{\tau}) = f(x;~U(\theta;~\mathcal{D}_{\tau}))
\end{equation}
We adopt the shorthand $ \Ls ( U (\theta) ) $  used in \cite{nichol2018reptile} to indicate that the loss $\Ls$ is computed over $f(x;~\theta_{\tau})$ for $ x, y \in \mathcal{D}_{\tau}$:
\begin{equation}
  \label{eq:LU}
  \Ls ( U (\theta) ) := \Ls ( f(x;~\theta_{\tau}), y )
\end{equation}
To minimize the loss incurred in the update routine, we first take the derivative with respect to the initialization
\begin{equation}
	\frac{\partial}{\partial \theta} \Ls ( U (\theta) ) = U'(\theta) \cdot \Ls'(U(\theta)) \label{eq:gg}
\end{equation}
where the resulting term  $U^{\prime}$ is the derivative of a gradient based update procedure, and hence, contains second order derivatives.
In first-order renditions explored in \cite{nichol2018reptile}, FOMAML and Reptile, finite differences are used to approximate the gradient of the meta-update $\nabla \theta$. The difference between the two approximations can be summarized by how they make use of $\mathcal{D}_{\tau}^{tr}$ and $\mathcal{D}_{\tau}^{val}$:
\begin{align}
  \label{eq:update}
  {\theta^{tr}} ~~ &\leftarrow U(\theta \quad ;  \mathcal{D}_{\tau}^{tr} ~, \omega^{tr} ~~) \\
  {\theta^{val}} ~ &\leftarrow U(\theta^{tr} ~; \mathcal{D}_{\tau}^{val}, \omega^{val}) \\
  {\theta^{both}} &\leftarrow U(\theta \quad ;  \mathcal{D}_{\tau}^{tr} \cup \mathcal{D}_{\tau}^{val} , \omega^{tr})
 \end{align}
 Reptile trains jointly on both, while FOMAML trains on the two sets separately in sequence, favoring initializations that differ less between the splits.
  \begin{align}
    \text{Reptile:} ~ \nabla \theta &\propto \theta^{both} - \theta \\
    \text{FOMAML:} ~ \nabla \theta &\propto \theta^{val} ~ - \theta^{tr} \label{eq:finite_difference}
  \end{align}
The gradient approximation $\nabla \theta$ can then be used to optimize the initialization by stochastic gradient descent or any other gradient-based update procedure.

\subsection{Optimizing Test-Time Update Hyperparameters}
\label{motivating_UHO}
As shown clearly in equations \ref{eq:f} and \ref{eq:LU}, the error of any function $f$ learned or predicted from a dataset $\mathcal{D}_{\tau}$ depends on the learning algorithm $U$. This analysis motivates research question 2 in section \ref{Introduction}, which asks how significant is the effect of the hyperparameters of $U$. To address this question, 
we leverage the flexibility to choose hyperparameters $\omega^{test}$ when adapting to new tasks, separately from the hyperparameters $\omega^{tr}$ used in meta-training.
The optimal choice of $\omega^{test}$ can be determined by minimizing the expected loss in eq. \ref{eq:1} with respect to the hyperparameters, treating $\hat{\theta}^*$ and $\mathcal{D}^{tr}_{\tau}$ as parameters of the update routine:
\begin{equation}
    \label{eq:motivating_UHO}
    \hat{\omega}^{*} = \argmin_{\omega} \E_{\hat{p}} \left[ \E_{\hat{q}_{\tau}} \left[ \Ls \left( U (  \omega ~; \hat{\theta}^{*}, \mathcal{D}^{tr}_{\tau})  \right) \right] \right]
\end{equation}
Empirical estimations of the optimal initialization $\hat{\theta}^*$ have an implicit dependence on $\mathcal{T}^{tr}$ and $\omega^{tr}$ (eq. \ref{eq:empirical_objective}), and the optimal hyperparameters $\hat{\omega}^{*}$ depend on the $\hat{\theta}^*$ in turn (eq. \ref{eq:motivating_UHO}). We call the general procedure of optimizing the update routine's hyperparameters to decrease meta-test-time error update hyperparameter optimization (UHO).

\section{EfficientLab Architecture for Image Segmentation}
To extend first-order MAML-type algorithms to more expressive models, with larger hypothesis spaces, while yielding state of the art few-shot learning results, we developed a novel neural network architecture, which we term EfficientLab. The top level hierarchy of the network's organization of computational layers is similar to \cite{chen2018encoder} with convolutional blocks that successively halve the features in spatial resolution while increasing the number of feature maps. This is followed by bilinear upsampling of features which are concatenated with features from long skip connections from the downsampling blocks in the encoding part of the network. The concatenated low and high resolution features are then fed through a novel atrous spatial pyramid pooling (ASPP) module, which we call a residual skip decoder (RSD), and finally bilinearly upsampled to the original image size.

\begin{figure}[!t]
\begin{center}
   \includegraphics[height=0.5\linewidth]{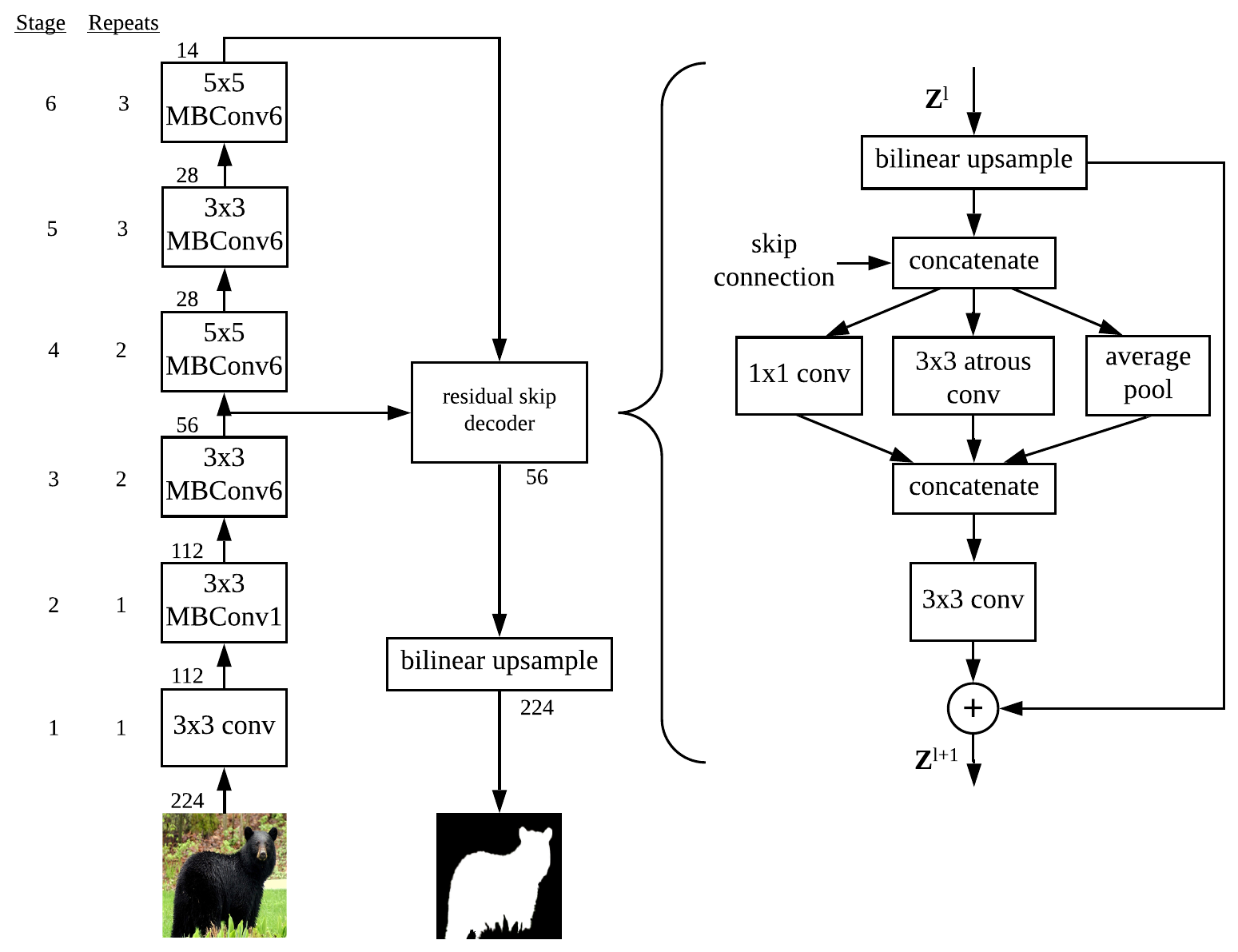}
\end{center}
   \caption{Diagram of the computations performed by the EfficientLab-3 neural network. Nodes represent functions and edges represent output tensors. Output spatial resolutions are written next to the output edge. The high level architecture shows the EfficientNet feature extractor on the left with mobile inverted bottleneck convolutional blocks (see~\cite{tan2019efficientnet,sandler2018mobilenetv2} for more details). On the right is the residual skip decoder module that we utilize in the upsampling branch of EfficientLab. The numbers suffixing EfficientLab denote the stage at which a residual skip decoder (RSD) is located, with EfficientLab-3, pictured, having an RSD with skip connections from stage 3 in the downsampling layers. EfficientLab-6-3 has RSD modules at both stages 6 and 3.}
\label{fig:rsd}
\end{figure}

The differences between our model and the DeepLab model are in (1) the encoder network used and (2) how the low resolution embedded features are upsampled to full resolution predictions. For the encoding subnetwork, we utilize the recently proposed EfficientNet~\cite{tan2019efficientnet}.
After encoding the images, the feature maps are upsampled through a parameterized number of RSD modules. The RSD computational graph of operations is shown in Figure \ref{fig:rsd}. EfficientLab-3 has one RSD module at the third stage while EfficientLab-6-3 has RSD modules at the 6th and 3rd stages as shown in figure \ref{fig:rsd}. The RSD module utilizes three parallel branches of a $1 \times 1$ convolution, $3 \times 3$ convolution with dilation rate = 2, and a simple average-pooling across spatial dimensions of the feature maps. The output of the three branches is concatenated and fed into a final $3 \times 3$ convolutional layer with 112 filters. A residual connection wraps around the convolutional layers to ease gradient flow \footnote{Residual connections have been suggested to make the loss landscape of deep neural networks more convex~\cite{li2018visualizing}. If this is the case, it could be especially helpful in finding low-error minima via gradient-based update routines such as those used by MAML, FOMAML, and Reptile.}. Before the final $1 \times 1$ convolution that produces the unnormalized heatmap of class scores, we use a single layer of dropout with a drop rate probability = 0.2 \footnote{As described in \cite{li2019understanding} and used in \cite{tan2019efficientnet} the dropout layer is applied after all batch norm layers.}. We use the standard softmax to produce the normalized predicted probabilities.

We use batch normalization layers following convolutional layers~\cite{ioffe2015batch}. We meta-learn the $\beta$ and $\gamma$ parameters, adapt them at test time to test tasks, and use running averages as estimates for the population mean and variance, $E[x]$ and $Var[x]$, at inference time as suggested in \cite{antoniou2018train}. \textit{All} parameters at the end of an evaluation call are reset to their pre-adaptation values to stop information leakage between the training and validation sets. The network is trained with the binary cross entropy minus the log of the dice score~\cite{dice1945measures}
, which we adapt from the loss function of~\cite{iglovikov2017satellite}, plus an $L_{2}$ regularization on the weights:
\begin{equation}
\mathcal{L} = H - log(J) + \lambda \left\|\theta\right\|_{2}^{2}
\end{equation}
where $H$ is binary cross entropy loss:
\begin{equation}
H=-\frac{1}{n} \sum_{i=1}^{n}\left(y_{i} \log \hat{y}_{i}+\left(1-y_{i}\right) \log \left(1-\hat{y}_{i}\right)\right)
\end{equation}
$J$ is the modified Dice score:
\begin{equation}
J = \frac{2IoU}{IoU + 1}
\end{equation}
and $IoU$ is the intersection over union metric:
\begin{equation}
IoU=\frac{1}{n} \sum_{i=1}^{n}\left(\frac{y_{i} \hat{y}_{i} + \epsilon}{y_{i}+\hat{y}_{i}-y_{i} \hat{y}_{i} + \epsilon }\right) \label{eq:IoU}
\end{equation}

\section{Experiments}
\label{experiments}
We evaluate the FOMAML and Reptile meta-learning algorithms on the FSS-1000 dataset. Model topology development

and meta-training hyperparameter search was done on the held out set of validation tasks, $\mathcal{T}^{val}$, and not the final test tasks. For the final evaluations reported in Table \ref{table:fss}, we meta-train for 50,000 meta-iterations, which is $\sim$330 epochs through the training and validation tasks $\mathcal{T}^{tr} \cup \mathcal{T}^{val}$ of the FSS-1000 dataset using a meta-batch size of 5, an inner loop batch size of 8, and 5 inner loop iterations. For reptile, we experiment with setting the train shots to 5 and 10. Note that when setting train shots to 5, this effectively causes the number of epochs through all individual image-pair examples to be cut in half, to $\sim$165 epochs through the individual examples. During training, we use stochastic gradient descent (SGD) in the inner loop with a fixed learning rate of 0.005. During training and evaluation, we apply simple augmentations to the few-shot examples including random translation, rotation, horizontal flips, additive Gaussian noise, brightness, and random eraser~\cite{zhong2017random}. We use $L_2$ regularization on all weights with a coefficient $\lambda = 5\mathrm{e}{-4}$.

\subsection{FSS-1000 Dataset}
\label{data}
The first few-shot image segmentation dataset was the PASCAL-5\textsuperscript{i} presented in \cite{shaban2017one} which reimagines the PASCAL dataset~\cite{everingham2010pascal} as a few-shot binary segmentation problem for each of the classes in the original dataset. Unfortunately, the dataset contains relatively few distinct tasks (20 excluding background and unlabeled). The idea of a meta-learning dataset for image segmentation was further developed with the recently introduced FSS-1000 dataset, which contains 1000 classes, 240 of which are dedicated to the meta-test set $\mathcal{T}^{test}$, with 10 image-mask pairs for each class~\cite{wei2019fss}. For each of the rows in the results table \ref{table:fss}, we evaluate the network on the 240 test tasks, sampling two random splits into training and testing sets for each task, yielding 480 data points per meta-learning approach for which the mean intersection over union (IoU) (eq. \ref{eq:IoU}) and 95\% confidence interval are reported. The FSS-1000 dataset is the focus of the empirical comparisons of network ablations and meta-learning approaches that we experiment with in this paper.

\subsection{FP-k Dataset}
For investigating how the meta-learned representations integrate new information as more data becomes a available, we put together a small benchmark dataset that we call FP-k. FP-k takes 5 tasks from FSS-1000 and 5 tasks from PASCAL-5\textsuperscript{i} for the same concept\footnote{See the supplementary material for more details on the dataset construction.}. Using this dataset, we train over a range of ``k''-training shots from ImageNet-trained \footnote{The encoder is trained on ImageNet, while the residual skip decoder and final layer weights are initialized in the same way as EfficientNet~\cite{tan2019efficientnet}}, joint-trained, and our meta-learned initializations.
 We report the performance of our EfficientLab network meta-trained with FOMAML over a range of k examples as a benchmark which we hope will inspire future empirical research into studying how meta-learning approaches scale in accuracy and computational complexity as more labeled data become available. For all three initializations, we use UHO for estimating the hyperparameters of $U$ for $k < 10$. For $k \ge 10$, we use a fixed learning rate and early stopping to estimate the optimal number of iterations. These results are shown in Figure~\ref{fig:k-shot-lr-curves} and discussed in \ref{results}.

\subsection{Test-Time Update Hyperparameter Optimization Methodology}
\label{hyperparameters}

Generalization in meta-learning requires both the ability to learn representations for new tasks efficiently ($\mathcal{T}^{tr}$ to $\mathcal{T}^{test}$), and to select representations that are able to capture unseen test examples effectively ($\mathcal{D}_{\tau}^{tr}$ to $\mathcal{D}_{\tau}^{test}$). The approximation scheme of FOMAML addresses the latter by taking the finite difference between updates using the train and validation sets (as shown in eq. \ref{eq:finite_difference}), favoring initializations that differ less between splits of $\mathcal{D}_{\tau}^{tr} \cup \mathcal{D}_{\tau}^{val}$.
In investigation of research question 2 in section \ref{Introduction} and to further improve generalization within task to $\mathcal{D}_{\tau}^{test}$, we tune $\omega$ after meta-learning $\hat{\theta}^{*}$ to find $\omega^{test}$ (as shown in eq. \ref{eq:motivating_UHO}). We use $\omega^{test}$ at meta-test time when adapting the initialization to new tasks. Specifically, we use Bayesian optimization with Gaussian processes to optimize the hyperparameters $\omega$ \cite{snoek2012practical}.
We apply this UHO procedure to estimate the optimal adaption routine's hyperparameters using 200 randomly validation tasks $\mathcal{T}^{val}$ that are held our from meta-training. We specifically search over the learning rate and the number of gradient updates that are applied when adapting to a new task $\tau$. We report results with and without optimized update hyperparameters in table \ref{table:fss}. We find that optimizing $\omega$ significantly improves adaptation performance on the meta-test tasks $\mathcal{T}^{test}$.

\subsection{Joint-Trained Initialization}
\label{joint}
We trained EfficientLab on the ``joint'' distribution of $\mathcal{T}^{tr} \cup \mathcal{T}^{val}$ in a standard training loop, without an inner loop. Each batch contained a random sample of examples from any of the classes as is standard in SGD. The only change to the network architecture was that instead of predicting 2 channel output (foreground and background), the network was trained to predict the number of task classes plus a background class. Other than these changes, we matched meta-training hyperparameters as faithfully as possible: training for 200 epochs, batch size of 8 image-mask pairs, using a learning rate of 0.005 with a linear decay, and regularization.

\subsection{Results}
\label{results}

We show the results of experimenting with different decoder architectures for EfficientLab in Table~\ref{table:EfficientLabAblations}.
Each network topology is meta-trained with FOMAML and the same meta-training hyperparameters defined in~\ref{experiments}. The bulk of our experiments were done with EfficientLab-3, including all FP-k experiments, though our best results were found using EfficientLab-6-3. We expect this trend of asymptotic improvements when increasing model dimensionality as also found in \cite{kaplan2020scaling}.

\begin{table}[!ht]
  \begin{center}
  \begin{tabular}{l c}
\multicolumn{1}{c}{\textbf{Network Architecture}} & \multicolumn{1}{c}{$\mathbf{\widebar{IoU}}$} \\
\hline
Auto-DeepLab decoder & $71.16 \pm 1.03\%$ \\
RSD at Stage 3 w/o residual & $77.55 \pm 1.08\%$ \\
RSD at Stage 3 & $79.89 \pm 0.98\%$ \\
RSD at Stages 6 \& 3 & $80.43 \pm 0.91\%$\\
\end{tabular}
\caption{{\bf EfficientLab architecture ablations}. Each network is meta-trained in the same way following Section \ref{experiments} and tested on the set of test tasks from FSS-1000~\cite{wei2019fss}. The row ``RSD at Stage 3 w/o residual'' contains results of removing the short-range residual connection from our proposed RSD module. The final row is the best network we find for few-shot performance via model agnostic meta-learning. We call this network EfficientLab in reference to the encoder of EfficientNet~\cite{tan2019efficientnet} and the decoder of DeepLab~\cite{chen2018encoder}, which it is inspired by.} Specifically, we use EfficientLab-3 to indicating that the RSD module is placed on the upsampling path at stage 3 and EfficientLab-6-3 to indicate that RSD modules are located on the upsampling path at both stages 6 and 3.
\label{table:EfficientLabAblations}
\end{center}
\end{table}

The results of our model with an initialization meta-learned using Reptile and FOMAML are shown in Table ~\ref{table:fss}. We find that EfficientLab trained with FOMAML and importantly with an adaptation routine optimized for low out of distribution test error, regularization, and improved use of batch normalization yields state of the art results.  Given that previous works have used regularization minimally or not at all during meta-training, we also conducted an ablation of removing regularization on the model. We find, unsurprisingly, that the combination of an $L2$ loss on the weights, with simple augmentations, and a final layer of dropout significantly increases generalization performance. After optimizing the update routine's hyperparameters, our approach sets the new state of the art for the FSS-1000 dataset. We have included a visualization of example predictions for a small set of randomly sampled test tasks in~\ref{fig:example-predictions}. See the supplementary material for additional examples and failure cases.

\begin{table*}[ht]
  {\small
  \centering
  \caption{{\bf Training paradigms}. Mean IoU scores of the EfficientLab-3 and EfficientLab-6-3 network evaluated on FSS-1000 test set of tasks for 1-shot and 5-shot learning. We report the FSS-1000 baseline from \cite{wei2019fss}. Our best found model combined FOMAML\textsuperscript{*}, EfficientLab-6-3, regularization, and UHO. FOMAML $D^{tr} \cap D^{val} = \varnothing$ denotes the original definition of FOMAML put forth in \cite{nichol2018reptile} in which $D^{tr}$ and $D^{val}$ are completely disjoint while FOMAML\textsuperscript{*} denotes that the two mini-datasets have been sampled with replacement from $D^{tr} \cup D^{val}$.}
  \subfloat[FSS-1000 1-shot]{
    \begin{tabular}{l c}
      \multicolumn{1}{c}{\textbf{Method}} & \multicolumn{1}{c}{$\mathbf{\widebar{IoU}}$} \\
      \hline
      FSS-1000 Baseline & 73.47\% \\
      ImageNet-trained encoder & $42.46 \pm 1.40 \%$ \\
      Joint-trained & $28.07 \pm 0.99$ \\
      Joint-trained + UHO & $32.07 \pm 1.17\%$ \\
      Reptile & $73.99 \pm 1.38$ \\
      FOMAML\textsuperscript{*} & $75.87 \pm 1.10\%$ \\
      \textbf{FOMAML\textsuperscript{*} + UHO} & $\mathbf{76.45 \pm 1.16}\%$ \\
    \end{tabular}
  }

  \subfloat[FSS-1000 5-shot]{
    \begin{tabular}{l c}
      \multicolumn{1}{c}{\textbf{Method}} & \multicolumn{1}{c}{$\mathbf{\widebar{IoU}}$} \\
      \hline
      FSS-1000 Baseline & $80.12\%$ \\
      ImageNet-trained encoder & $50.26 \pm 1.41\%$ \\
      Joint-trained on FSS-1000 & $25.03 \pm 0.94\% $\\
      Joint-trained on FSS-1000 + UHO & $45.05 \pm  1.37\%$ \\
      Reptile & $79.78 \pm 0.95\%$\\
      FOMAML $D^{tr} \cap D^{val} = \varnothing$ & $75.02 \pm 1.07\%$ \\
      FOMAML\textsuperscript{*} - regularization & $77.89 \pm 1.03\%$ \\
      FOMAML\textsuperscript{*} & $79.89 \pm 0.98\%$ \\
      FOMAML\textsuperscript{*} + UHO & $81.36 \pm 0.80\%$ \\
      \textbf{EffLab-6-3 FOMAML\textsuperscript{*} + UHO} & $\mathbf{82.78 \pm 0.74}\%$ \\
      \label{table:fss}
    \end{tabular}
  }}
\end{table*}

\begin{figure}[htp]
  \centering
    \subfloat{\includegraphics[width=0.24\linewidth]{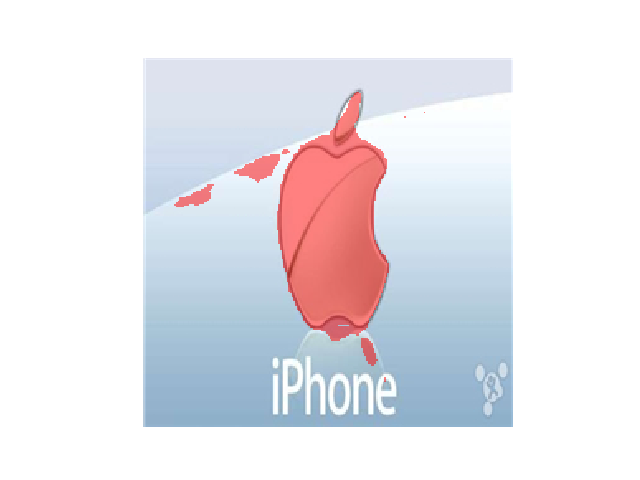}}
    \subfloat{\includegraphics[width=0.24\linewidth]{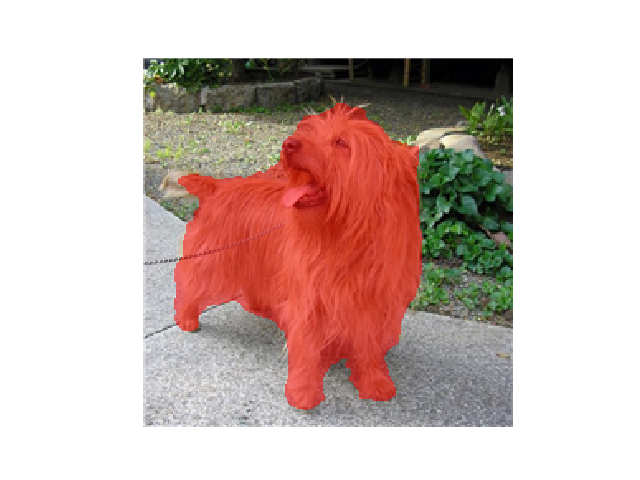}}
    \subfloat{\includegraphics[width=0.24\linewidth]{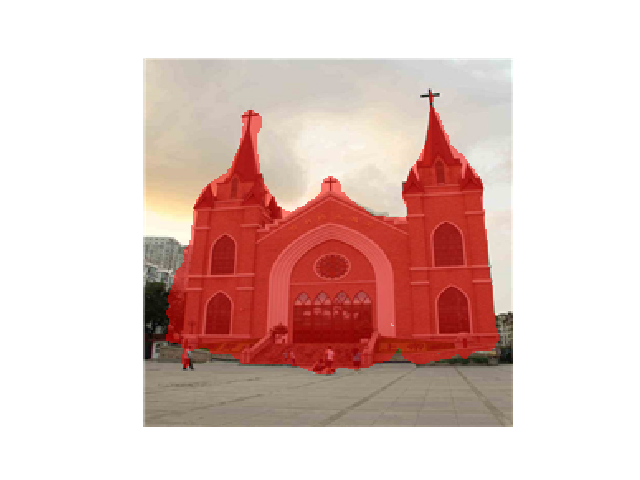}}
    \subfloat{\includegraphics[width=0.24\linewidth]{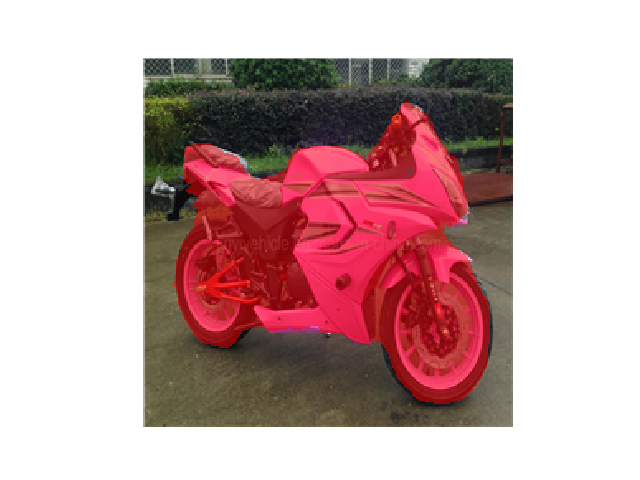}} \par
    \setcounter{subfigure}{0}
    \subfloat{\includegraphics[width=0.24\linewidth]{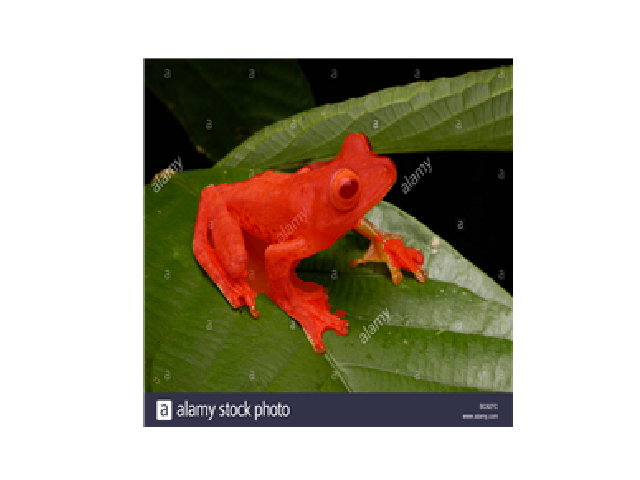}}
    \subfloat{\includegraphics[width=0.24\linewidth]{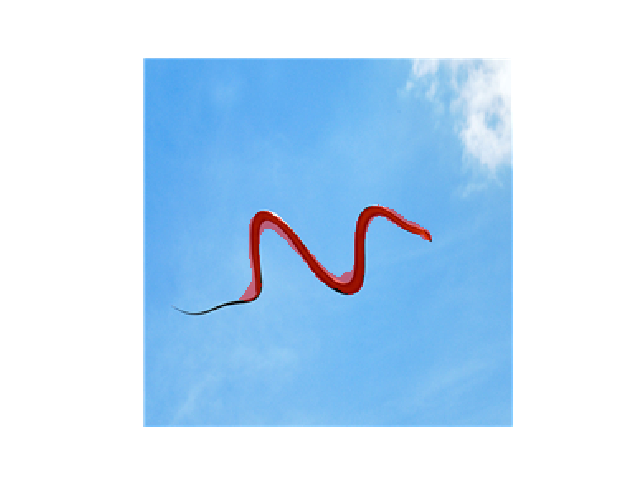}}
    \subfloat{\includegraphics[width=0.24\linewidth]{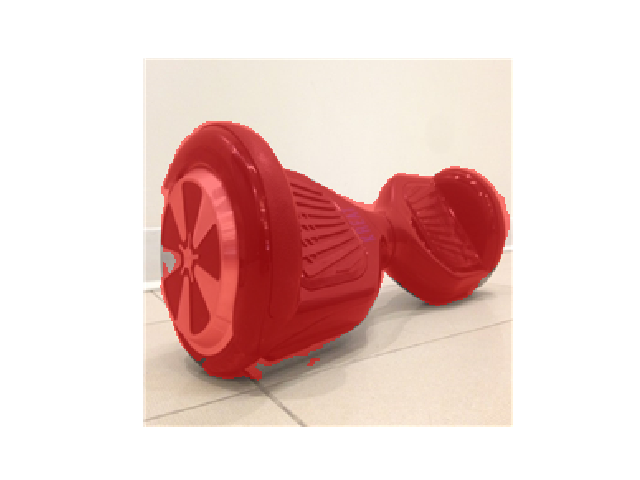}}
    \subfloat{\includegraphics[width=0.24\linewidth]{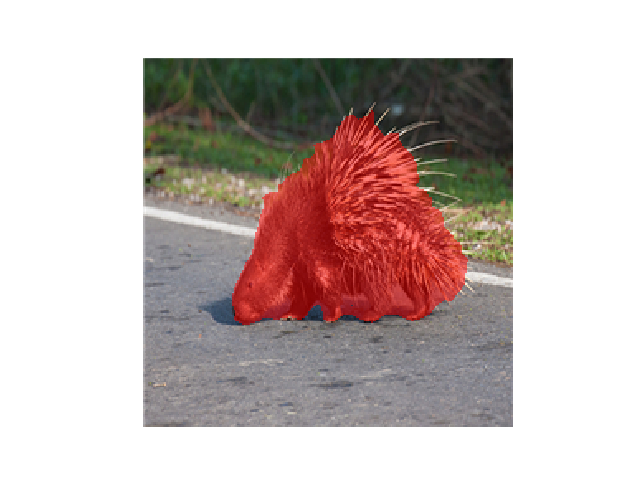}}
    \caption{Randomly sampled example 5-shot predictions on the test images from test tasks. Positive class prediction is overlaid in red. From left to right, top to bottom, the classes are \texttt{apple\_icon}, \texttt{australian\_terrier}, \texttt{church}, \texttt{motorbike}, \texttt{flying\_frog}, \texttt{flying\_snakes}, \texttt{hover\_board}, \texttt{porcupine}}
    \label{fig:example-predictions}
\end{figure}

Importantly, we also find that the original definition of FOMAML in which the mini-datasets $D^{tr}$ and $D^{val}$ are disjoint yields worse results than sampling with some amount of overlap. We find that by sampling $D^{tr}$ and $D^{val}$ with replacement from $D^{tr} \cup D^{val}$ yields better results. This sampling procedure is denoted by FOMAML\textsuperscript{*} below and can be interpreted as an interpolation between the original Reptile and FOMAML definitions put forth in \cite{nichol2018reptile}. We suspect that this sampling strategy serves as a form of meta-regularization though further work would be required on this detail to be conclusive. Similarly, we find that meta-training with Reptile using all 10 examples in each task of FSS-1000 produces worse results than meta-training with 5 examples per task in each meta-example.

\begin{figure}[!t]
\begin{center}
   \includegraphics[width=0.5\linewidth]{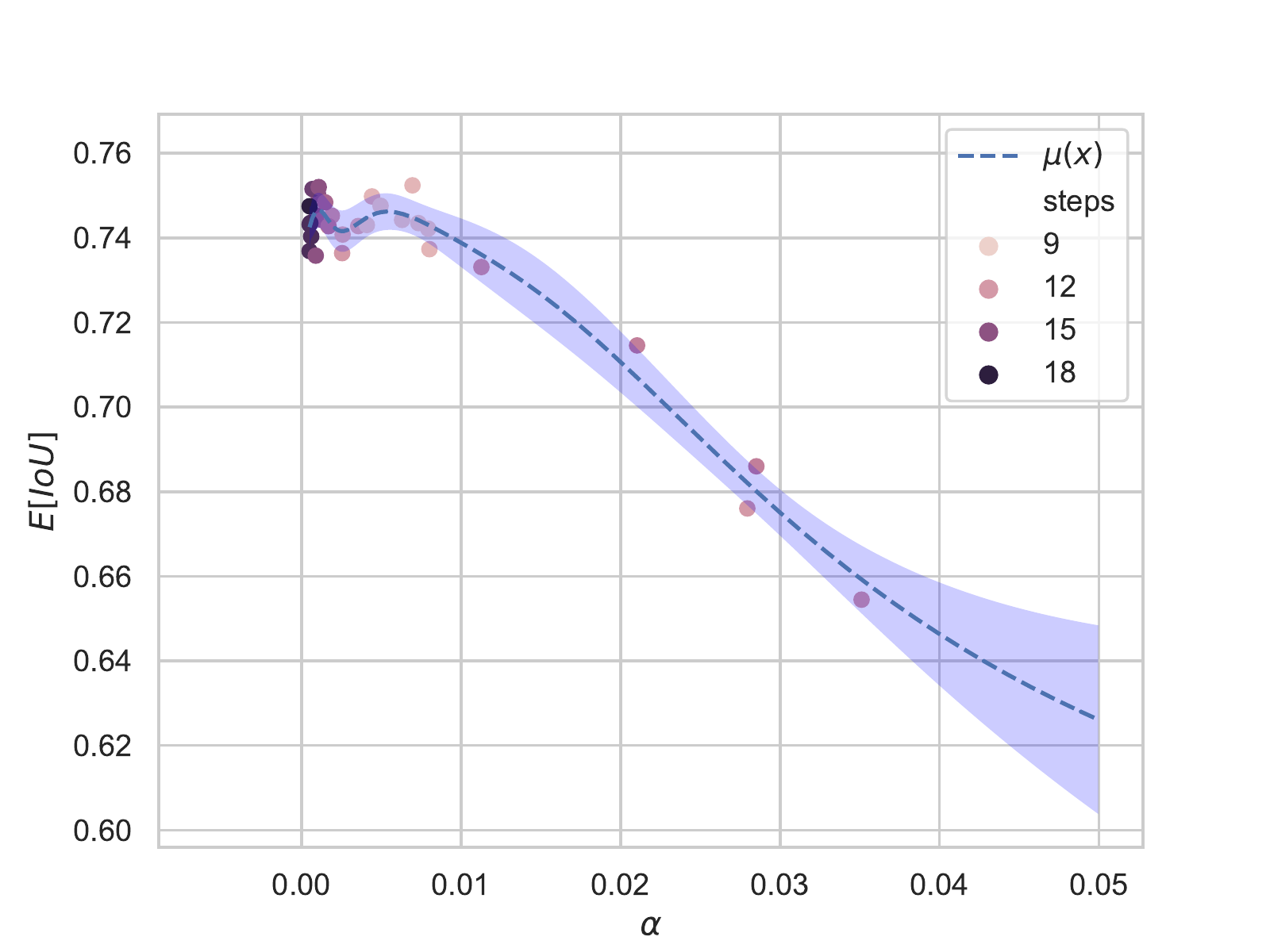}
\end{center}
   \caption{\textbf{Update hyperparameter optimization}. Mean IoU is shown as a function of the learning rate  $\alpha$ and the number of gradient steps over the set of 200 validation tasks $\mathcal{T}^{val}$. During optimization of the learning rate and the number of steps, the relationship between the learning rate and the IoU is modeled as a Gaussian process (shown in blue dashed line with 95\% confidence interval). Points are colored by median of the iterations each task was trained for before stopped by early stopping.}
\label{fig:iou-over-lr}
\end{figure}

To address research question 2 in section \ref{Introduction}, we also searched through a range of update routine learning rates, $\alpha$, that were $10\times$ less to $10\times$ greater than the learning rate used during meta-training. As clearly shown in Figure \ref{fig:iou-over-lr}, the learned representations are {\bf not} robust to such large variations in the hyperparameter.

In this work we posit that a fixed update procedure that is used at test time and not conditioned on the labeled examples for a new task $\mathcal{D}^{tr} \in \tau$
is one of the major hinderances of applying MAML-type algorithms to unseen tasks. In section \ref{prelim}, we show that the hyperparameters $\omega$ that are used to adapt the networks weights $\theta$ to a new task $\mathcal{D}_{\tau}$ can be optimized. We find this analysis to be supported empirically as well. We find that: (1) the estimated optimal hyperparameters for the update routine on the validation tasks are not the same as those specified a priori during meta-training, as illustrated in Figure \ref{fig:iou-over-lr}. One may expect that MAML-type algorithms would converge to a point in parameter space from which optimal minima for each of the training tasks are reachable. We find that even after 330 epochs through the training tasks, this was not the case. The best learning rate and number of iterations we found via UHO on the held-out validation set of 200 tasks $\mathcal{T}^{val}$ were orders of magnitude smaller for the learning rate and orders of magnitude greater for the best number of iterations respectively. We discuss this phenomena in more depth in the supplementary material. Furthermore, we noticed that the larger the number of maximum steps that UHO was allowed to search within, the smaller the optimal learning rate that was returned.
(2) Optimizing the hyperparameters after meta-training improves test-time results on unseen tasks. Furthermore, we find that \textit{meta-training} from scratch (and evaluating) with the UHO-selected hyperparameters yields nearly identical results to meta-training with the initial hyper parameters learning rate = 0.005 and inner-iterations = 5.
This further suggests that it may be useful to tune the hyperparameters $\omega$ \textit{after} meta-training to improve the generalization performance of the gradient-based adaptation routine $U$.

\begin{figure}[htp]
   \centering
     \subfloat{\includegraphics[width=0.33\linewidth]{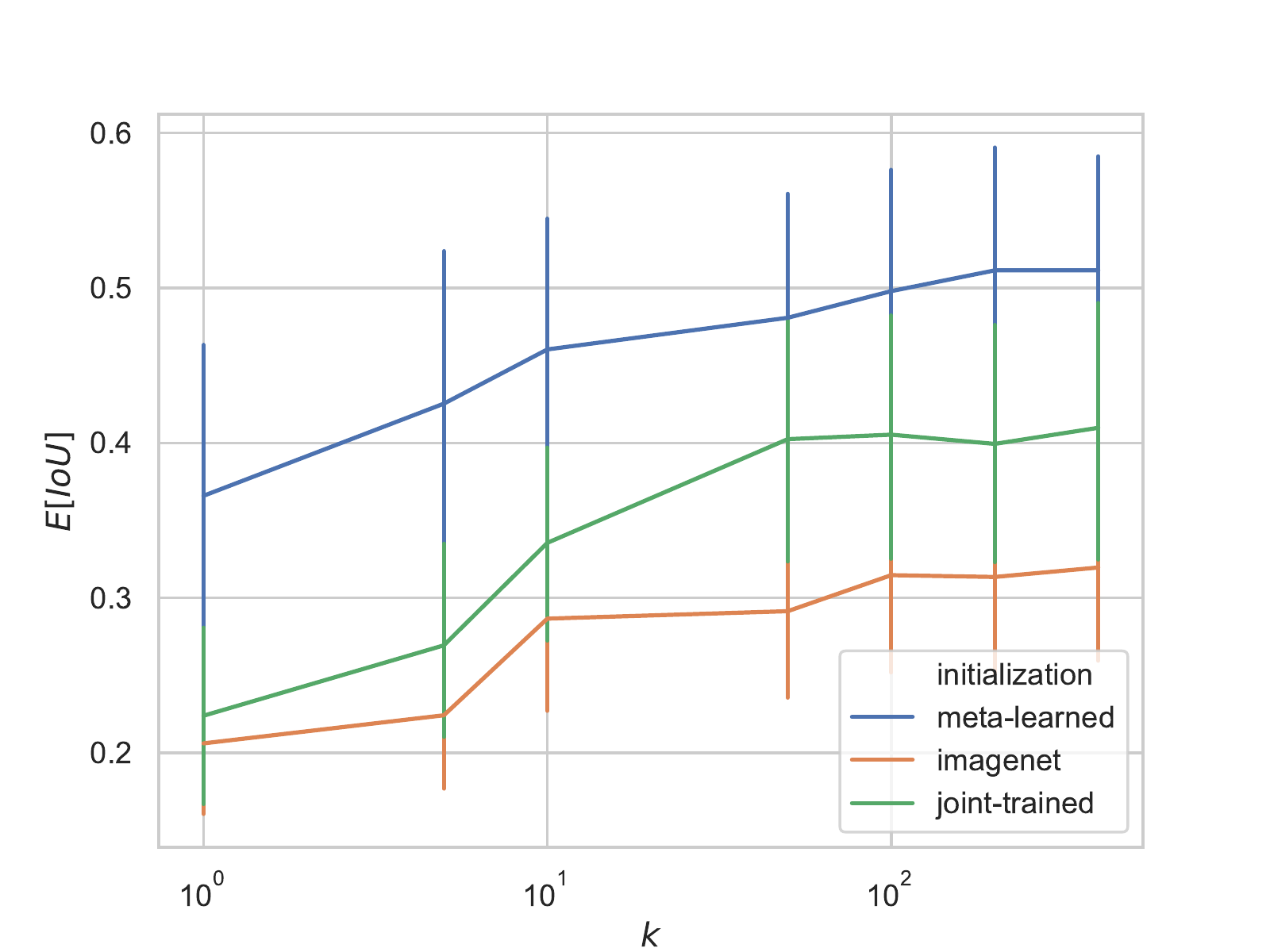}}
     \subfloat{\includegraphics[width=0.33\linewidth]{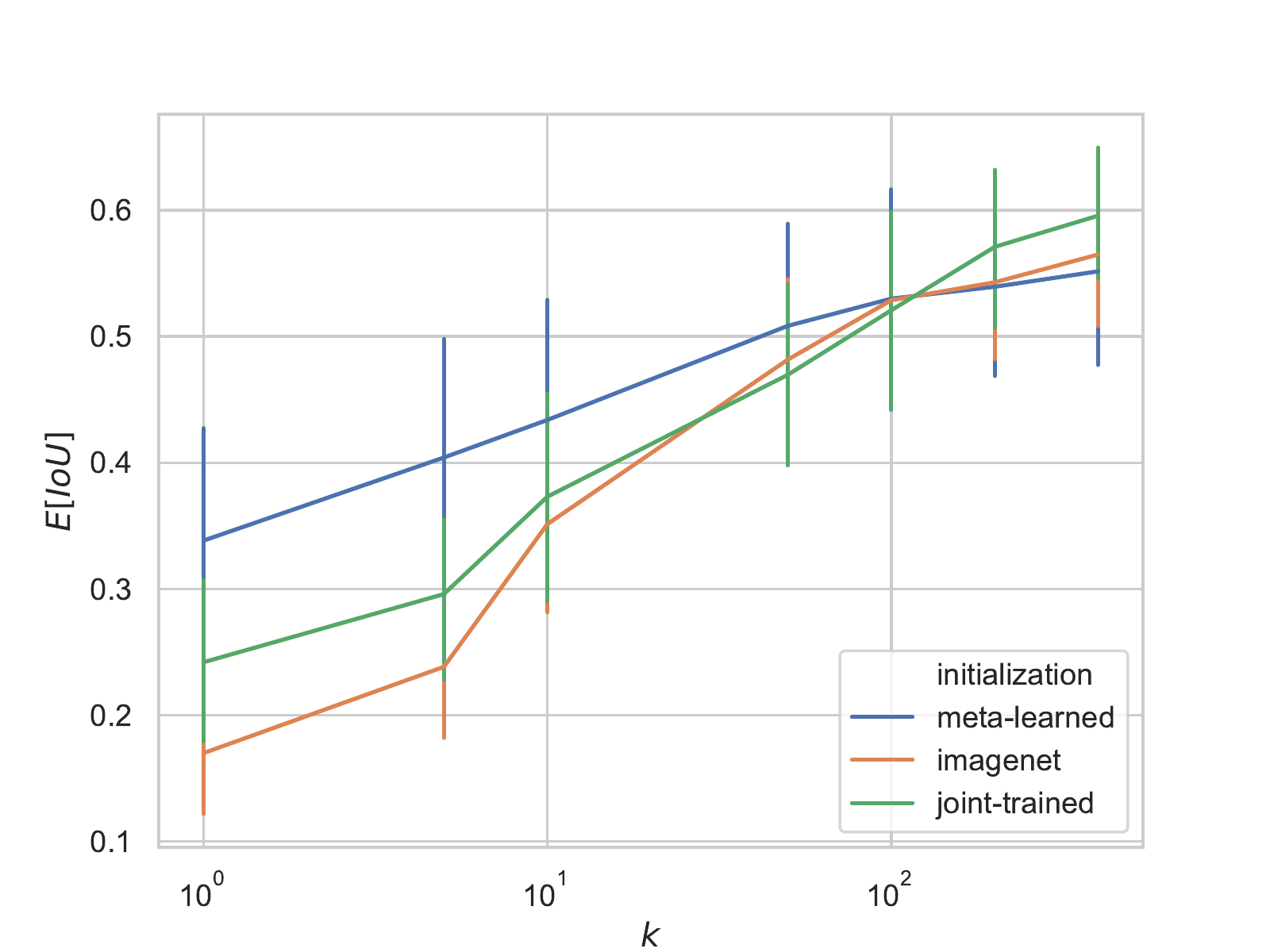}}
     \subfloat{\includegraphics[width=0.33\linewidth]{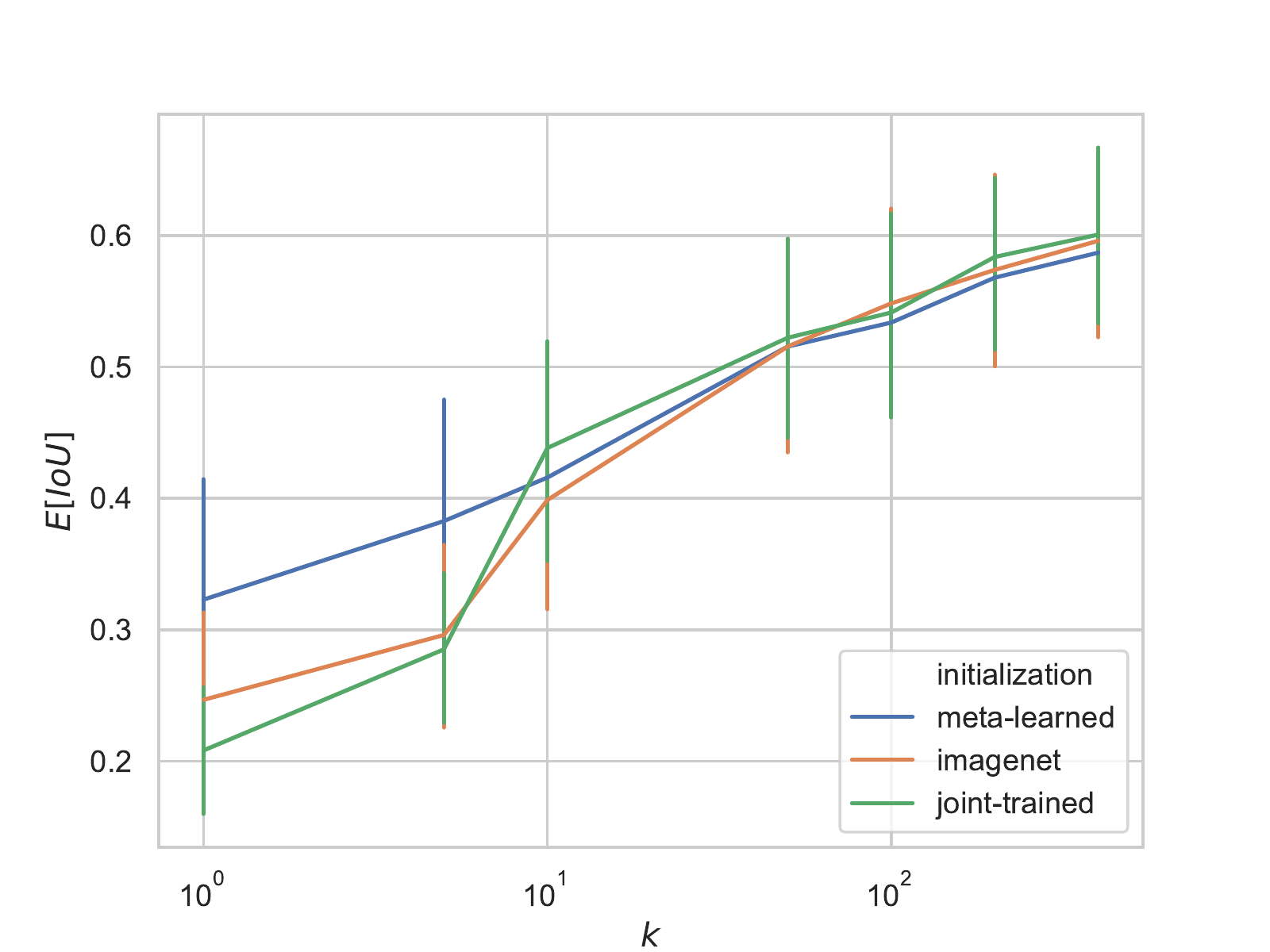}}
     \caption{Mean IoU results as a function of the training set size of our EfficientLab-3 model adapted to tasks of the FP-k dataset. Given the importance we have shown in the adaption routine's hyperparameters, we experiment with different learning rate schedules and early stopping parameters. On the left, are the results from using a step decay learning rate as in \cite{wilson2017marginal} with early stopping patience of 100 gradient steps. In the middle and right are the results from using a fixed learning rate but with an early stopping patience of 50 and 100 gradient steps respectively.}
      \label{fig:k-shot-lr-curves}
 \end{figure}

 By training our model on the FP-k dataset, we found that our meta-learned initializations reliably outperformed joint-trained, and ImageNet-trained initializations but only up to 10 training examples\footnote{The examples in the PASCAL dataset are known to be more challenging than the FSS-1000 dataset ~\cite{wei2019fss}. From visual inspection of the two datasets, it is also clear that the PASCAL dataset contains more label noise than the FSS-1000 dataset. For these reasons, the mean IoU values shown in Figure \ref{fig:k-shot-lr-curves}, which contain examples from both datasets, are not directly comparable to the results shown in Table \ref{table:fss}, which contain examples only from FSS-1000.}.
 As shown in Figure \ref{fig:k-shot-lr-curves}, we again find that the hyperparameters of $U$ are critical for effectively adapting the parameters to new tasks. In particular, we find that a large early stopping patience is necessary for maximal performance when adapting to new tasks as learning, especially with non-meta-learned initializations, undergoes a period where out of distribution error increases. We suspect that the larger patience is required for good generalization as the number of training examples grows due to the concept of double descent noted in \cite{nakkiran2019deep}.

\section{Future Work}
In future work, it would be useful to take a more critical look at the interplay between batch normalization and meta-learning. While single task deep neural networks in large data regimes apply batch normalization with a consistent pattern, different groups working in few-shot meta-learning have incorporated batch norm in markedly different ways such as by: (1) not using it at all for the meta-learning components~\cite{rusu2018meta}, (2) not using learned $\beta$ and $\gamma$ parameters at all while still using estimated population means and variances during inference ~\cite{zhang2019canet}, or (3) meta-learning $\beta$ and $\gamma$ while only using batch statistics for the normalization ~\cite{finn2017model,nichol2018reptile}, or (4) meta-learning $\beta$ and $\gamma$ and also using population estimates of the mean and variance, as done conventionally when training deep neural networks in the large data regime, which is the approach that we adopt and find to be most useful. Another interesting question to address would be to evaluate how EfficientLab performs on more standard many-shot multi-class image segmentation problems such as the CityScapes dataset ~\cite{cordts2016cityscapes}.

\section{Conclusions}
In this work, we showed that gradient-based first order model agnostic meta-learning algorithms extend to the high dimensionality of the hypothesis spaces and the skewed distributions of few-shot image segmentation problems, but are sensitive to the hyperparameters, $\omega$, of the update routine, $U$. Furthermore, we find that the initializations that are meta-learned quickly match test time performance of traditional transfer learning approaches as more data becomes available.
These results, while produced from a small number of tasks, provide important context on the value in terms of labeled data efficiency of applying FOMAML to image segmentation.
Future work should investigate more critically, both empirically and theoretically, the efficacy of few-shot learning systems as more labeled data becomes available. To this end, we have reported results on a novel meta-test benchmark dataset, FP-k, which contains 5 tasks with 400 training and 20 test examples per task.

We also showed that the optimal hyperparameter configuration for the test-time update routine may not be the same configuration used during meta-learning. These findings are supported by our theoretical analyses which show that MAML-type algorithms minimize the empirical risk on the training set of a fixed update routine and the initialization $\theta$, but do not natively guarantee that the update routine's hyperparameters are optimal. We suspect that improvements realized by relation networks \cite{wei2019fss,zhang2019canet,rusu2018meta}, models that learn to generate parameters conditioned on the training data \cite{rusu2018meta,shaban2017one}, and models with learned learning rates \cite{li2017meta,antoniou2018train} directly leverage information on \textit{how} to adapt given a few-shot sample of labeled examples. We also suspect that the previous work in \cite{mishra2018a} may have found MAML-type algorithms to overfit when applied to high dimensional parameter spaces due to lack of regularization and lack of an empirical risk minimization of the update routine's hyperparameters.
Lastly, we hope that this work draws, what we argue is necessary, attention to the open problem of building learning systems that can unify small and large data regimes by gaining expertise and integrating new information as more data becomes available, much as people do.

%
\bibliographystyle{unsrt}
\bibliography{references}

\newpage
\begin{appendices}

\newfloatcommand{capbtabbox}{table}[][\FBwidth]

\floatsetup[table]{capposition=top}

\date{\vspace{-5ex}}

\section{Meta-Training Details}
For meta-training, we adapted the code in \cite{openai2018supervised}. We referenced the hyperparameters used in the Reptile meta-training runs for Mini-ImageNet in \cite{nichol2018reptile}. Due to the computational cost of meta-training and the combinatorial expansion in the meta-training hyperparameter search space due to having effectively the same number of hyperparameters for both the outer and inner meta-training loops, we did not exhaustively search over all meta-training hyperparameters. We did initial experimentation on tuning the inner batch size and the number of inner-loop gradient steps by evaluating on the validation tasks $\mathcal{T}^{val}$, though found that fine-tuning these values mattered less than optimizing the test-time hyperparameters.

\begin{table}[!ht]
\begin{center}
  \begin{tabular}{l c c }
    \hline
    Hyperparameter & value\\
    \hline
     Meta-batch size & 5 \\
     Meta-steps & 50000 \\
     Initial meta-learning rate &  0.1 \\
     Final meta-learning rate &  $1.e -5$ \\
     Inner batch size & 8 \\
     Inner steps & 5 \\
     Inner learning rate & 0.005 \\
     Final layer dropout rate & 0.2 \\
     Augmentation rate & 0.5 \\
    \hline
  \end{tabular}
  \caption{Meta-training hyperparameters for Reptile and FOMAML algorithms.}
  \label{table:meta-training-hyper-params}
\end{center}
\end{table}

Both Reptile and FOMAML were meta-trained with the hyperparameters shown in Table \ref{table:meta-training-hyper-params}. The only hyperparameter that we found significantly changed the results between the two approaches was the use of the ``train-shots'', which is the number of examples that each inner batch can sample from. We found that setting the Reptile train-shots equal to 10, which is the total number of examples per task in the FSS-1000 dataset, significantly reduced test-time performance. By decreasing the train-shots to 5, mIoU increased by $\approx5\%$ absolute percentage points of mean intersection over union (IoU). Similarly, we found that if we sampled $D^{tr}$ and $D^{val}$ with replacement, as opposed to using all 10 train-shots for FOMAML (using 5 for $D^{tr}$ and 5 for $D{val}$), meta-test results increased by $\approx5\%$ absolute percentage points of mean IoU. We suspect that limiting the number of train shots in this way serves as a form of meta-regularization.

\section{Test-Time Update Hyperparameter Optimization Method Details}
\label{UHO}
Motivated by the insight discussed in the preliminaries that the loss of a meta-learning algorithm with a fixed initialization $\theta$ is a function of the update routine $U$ \textit{and} its hyperparameters $\omega$, we tune $\omega$ to maximize intersection over union (IoU) on out-of-distribution tasks $\mathcal{T}^{val}$. We call this procedure inference hyperparameter optimization (UHO). For UHO, we use Bayesian optimization (BO) with a Gaussian process (GP) estimator of the objective function which is a standard approach for optimizing expensive black box functions \cite{snoek2012practical}. For the joint-trained and meta-learned initializations evaluated in Table 2, we experimented with tuning their learning rate by evaluating 30 parameters on each of the 200 validation tasks. The first half of the parameters were randomly from a log-uniform distribution and the second half were sampled from the posterior of the GP to maximize expected improvement in mean IoU over the validation tasks. For all evaluations, we optimize the learning rate over the interval $[0.0005, 0.05]$.

Because the effects of the learning rate are intertwined with the number of gradient updates, we also leveraged early stopping to decrease runtime and to estimate the optimal number of gradient steps when adapting to a new task. The use of early stopping in this way is purely an runtime optimization that reduces the search space that is explored when tuning $\omega$. We train each validation set $\mathcal{T}^{val}$ task independently and record the optimal number of gradient updates for each task. We then evaluate all tasks in $\mathcal{T}^{val}$ at the median number of steps returned by early stopping across tasks in $\mathcal{T}^{val}$.  We could have also used the Bayesian optimization with GP prior, but early stopping has the advantage of computational efficiency. Early stopping has been deeply studied with strong empirical and theoretical evidence to support its efficacy as an efficient hyperparameter tuning algorithm \cite{prechelt1998early,bengio2012deep,goodfellow2016deep,oymak2019generalization}. To reduce computational expense, we restricted the maximum number of gradient steps to 20 for each task.

\begin{table}[!ht]
\begin{center}
  \begin{tabular}{l c c }
    \hline
    Initialization & learning rate & steps\\
    \hline
     Joint-trained 1-shot & $8.156 e -4$ & 10\\
     Joint-trained 5-shot & $1.364 e -3$ & 17\\
     FOMAML\textsuperscript{*}  1-shot & $ 1.734 e -3$ & 8\\
     FOMAML\textsuperscript{*} 5-shot & $6.951 e -3$ & 12\\
    \hline
  \end{tabular}
  \caption{Inference hyperparameters returned from BO with GP for initializations of the EfficientLab-3 network. All other hyperparameters were fixed to the values shown in Table \ref{table:meta-training-hyper-params}.}
  \label{meta-test-hyperparams}
\end{center}
\end{table}

For our largest model, EfficientLab-6-3, we also experimented with using BO on a larger set of hyperparameters and larger number of maximum iterations for early stopping. We search over the learning rate $[0.0005, 0.05]$, the final layer dropout rate $[0.2, 0.5]$, the augmentation rate $[0.5, 1.0]$, and batch size $[1, 10]$ with a maximum early stopping iterations of 80.

\begin{table}[!ht]
\begin{center}
  \begin{tabular}{l c c c c c}
    \hline
    Initialization & learning rate & steps & dropout rate & augmentation rate & batch size\\
    \hline
     FOMAML\textsuperscript{*} 5-shot & $5 e -4$ & 59 & 0.5 & 0.5 & 8\\
    \hline
  \end{tabular}
  \caption{Inference hyperparameters returned from BO with GP for the EfficientLab-6-3 network.}
  \label{meta-test-hyperparams}
\end{center}
\end{table}

\section{Example predictions}
\label{example-predictions}
\begin{figure}[H]
  \centering
  \subfloat{\includegraphics[width=0.2\linewidth]{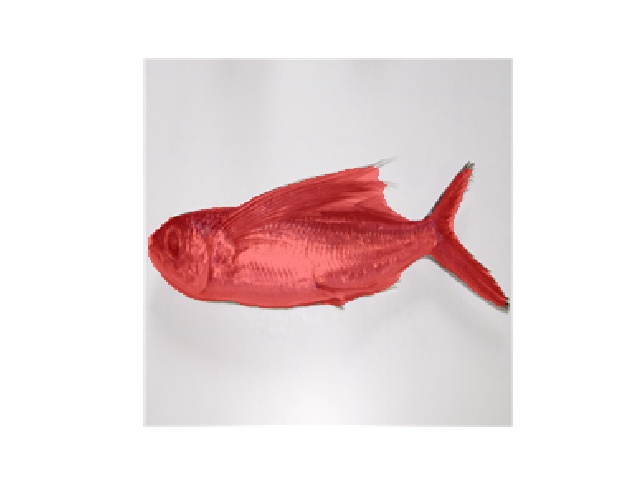}}
  \subfloat{\includegraphics[width=0.2\linewidth]{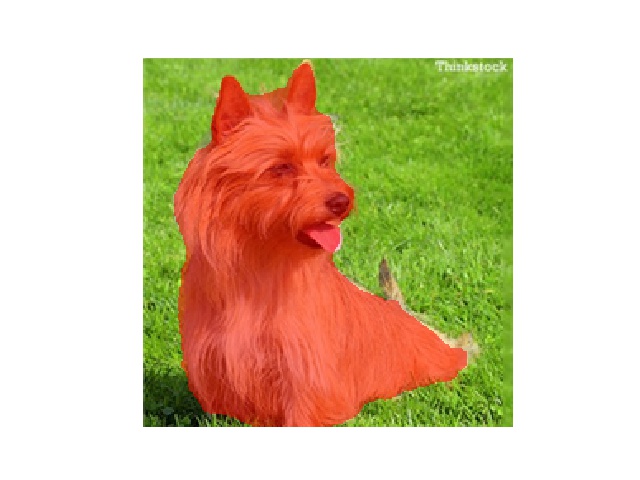}}
  \subfloat{\includegraphics[width=0.2\linewidth]{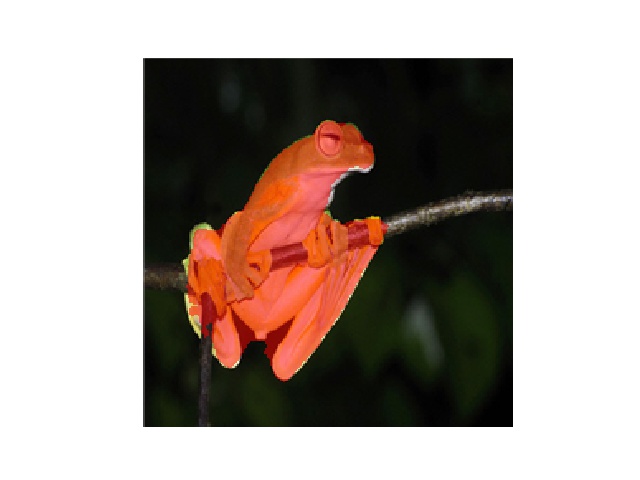}}
  \subfloat{\includegraphics[width=0.2\linewidth]{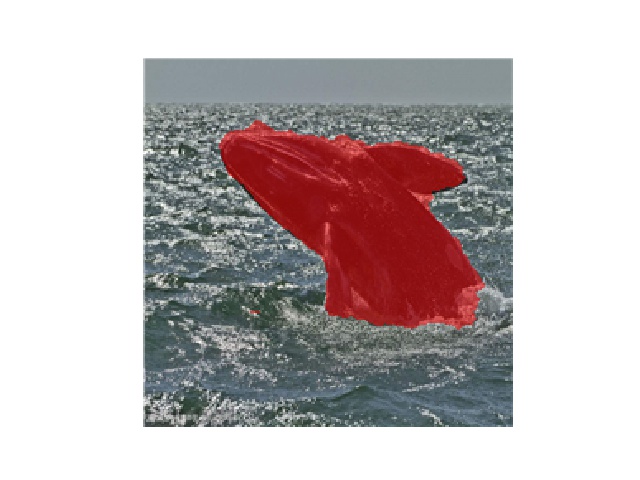}}
  \par
  \subfloat{\includegraphics[width=0.2\linewidth]{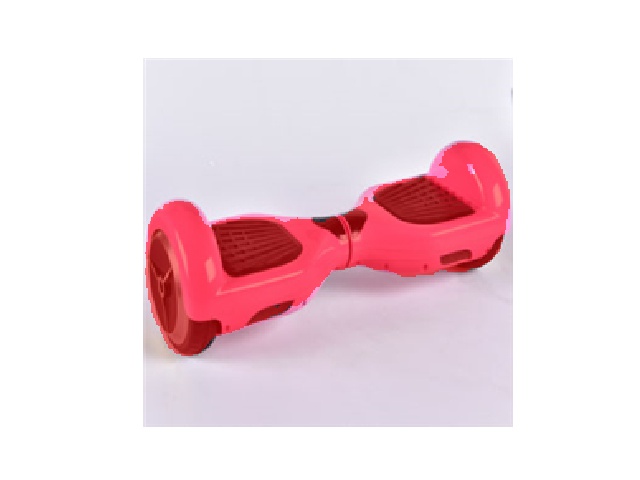}}
  \subfloat{\includegraphics[width=0.2\linewidth]{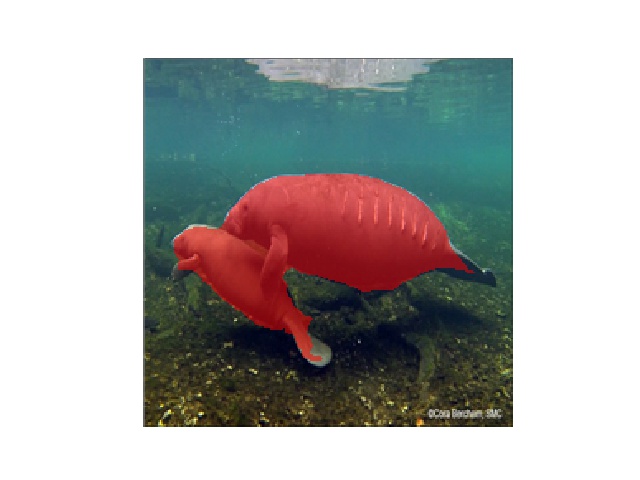}}
  \subfloat{\includegraphics[width=0.2\linewidth]{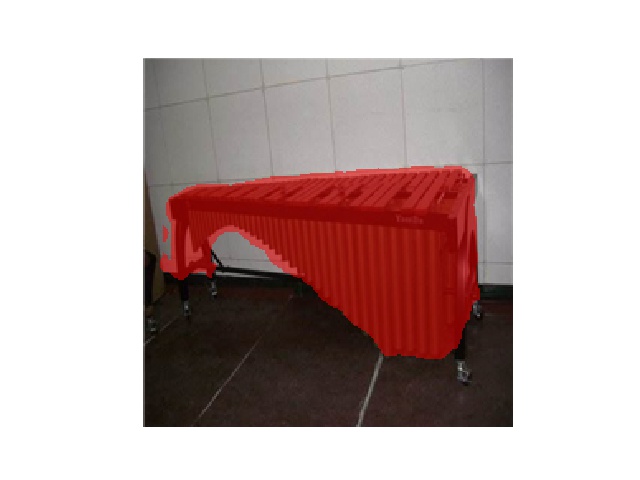}}
  \subfloat{\includegraphics[width=0.2\linewidth]{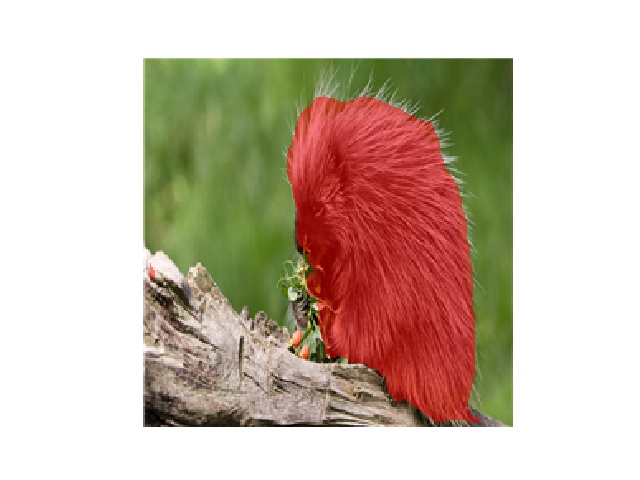}}
  \par
  \subfloat{\includegraphics[width=0.2\linewidth]{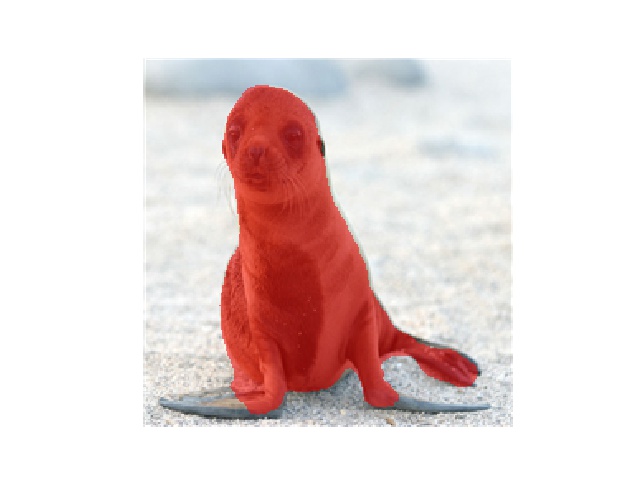}}
  \subfloat{\includegraphics[width=0.2\linewidth]{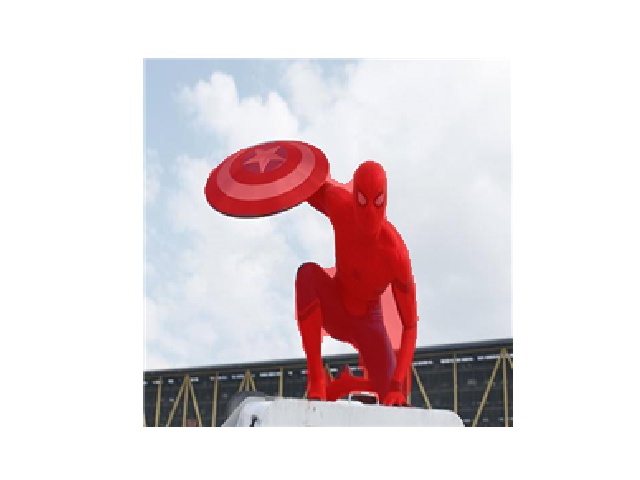}}
  \subfloat{\includegraphics[width=0.2\linewidth]{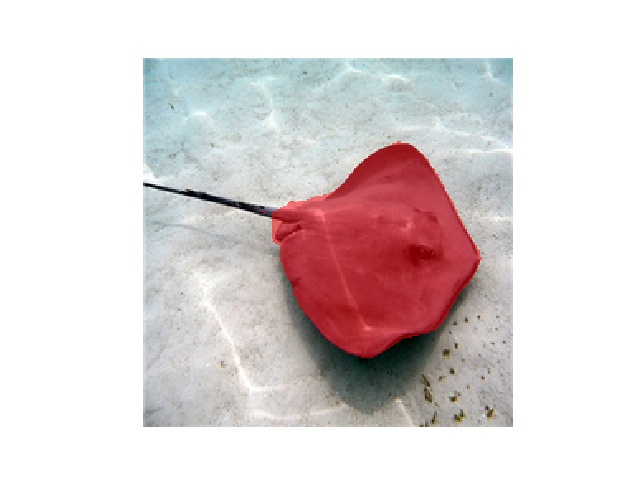}}
  \subfloat{\includegraphics[width=0.2\linewidth]{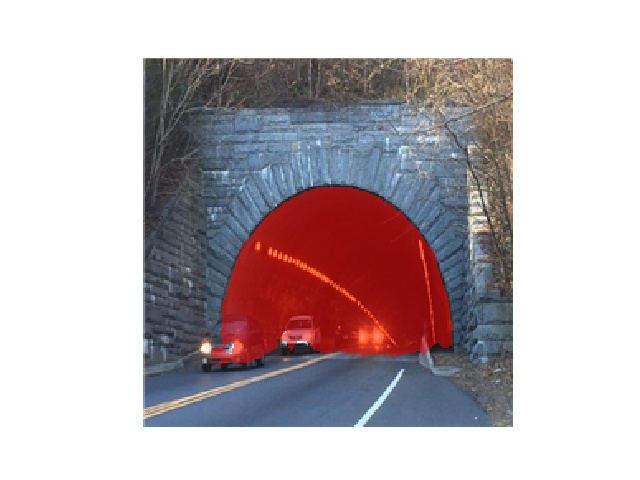}} \par
    \setcounter{subfigure}{0}
    \subfloat{\includegraphics[width=0.2\linewidth]{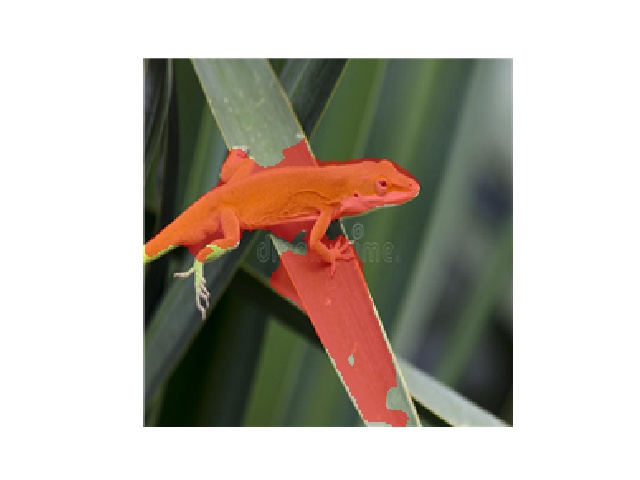}}
    \subfloat{\includegraphics[width=0.2\linewidth]{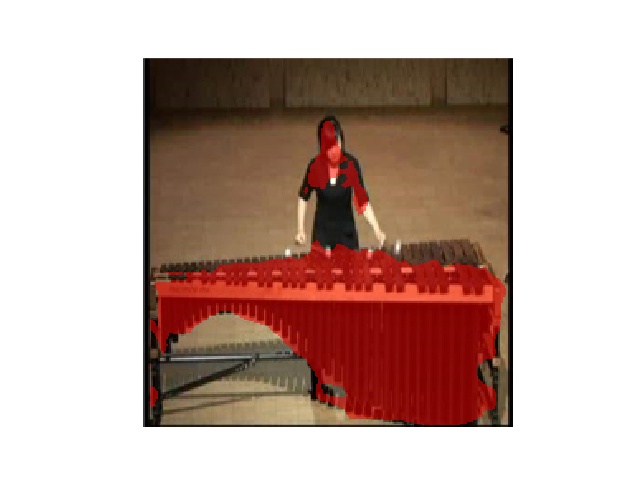}}
    \subfloat{\includegraphics[width=0.2\linewidth]{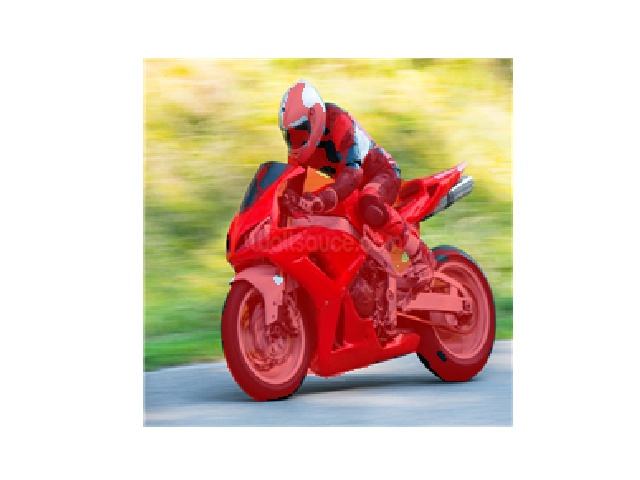}}
    \subfloat{\includegraphics[width=0.2\linewidth]{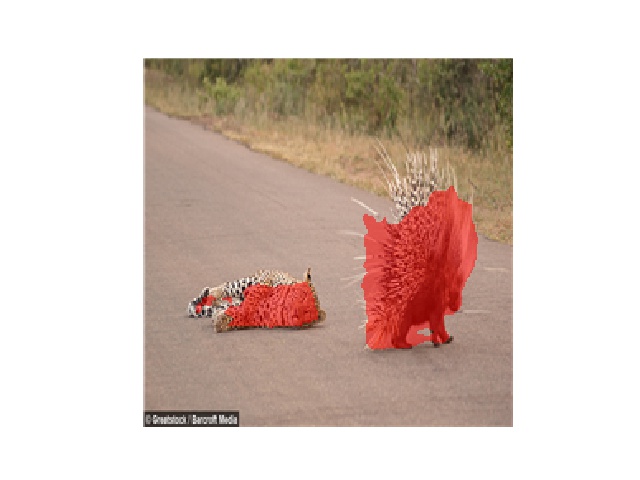}}
    \caption{Randomly sampled example 5-shot predictions on the test images from test tasks. Predictions were generated by EfficientLab-6-3 model meta-trained with FOMAML\textsuperscript{*} and evaluated with UHO-returned hyperparameters. Positive class prediction is overlaid in red. From left to right, top to bottom, the classes are \texttt{abes\_flying\_fish}, \texttt{australian\_terrier}, \texttt{flying\_frog}, \texttt{grey\_whale},  \texttt{hoverboard}, \texttt{manatee}, \texttt{marimba}, \texttt{porcupine}, \texttt{sealion}, \texttt{spiderman}, \texttt{stingray}, \texttt{tunnel}. The final row contains hand-picked failure cases from tasks \texttt{american\_chameleon}, \texttt{marimba}, \texttt{motorbike}, and \texttt{porcupine}.}
    \label{fig:ex-preds}
\end{figure}

We have included in Figure \ref{fig:ex-preds} a visualization of additional, randomly sampled predictions on test examples $\mathcal{D}^{test}$ from test tasks $\mathcal{T}^{test}$ that were never seen during meta-training. The failure cases are particularly interesting in that they suggest that a foreground object-ness prior has been learned in the meta-learned initialization.

\section{FP-k Dataset}
Table~\ref{table:pascal-5-tasks} contains the five tasks in PASCAL-5\textsuperscript{i} that have direct analogs in FSS-1000.
Each row contains the name of a task in FSS-1000 and PASCAL-5\textsuperscript{i}, respectively. We combine all examples for synonymous tasks. During evaluation, we simply randomly sample 20 test examples,
and sample a training set of $k$ examples over the range:
$ [1,~ 5,~ 10,~ 50,~ 100,~ 200,~ 400] $. For more details on our training and evaluation procedures see \ref{FP-k_Experimental_Details}.
\begin{table}[!ht]
\begin{center}
  \begin{tabular}{|c|c|}
    \hline
    PASCAL-5\textsuperscript{i} Task & FSS-1000 Task \\
    \hline
    \verb|aeroplane| & \verb|airliner| \\
    \verb|bus| & \verb|bus|\\
    \verb|motorbike| & \verb|motorbike| \\
    \verb|potted_plant| & \verb|potted_plant| \\
    \verb|tvmonitor| & \verb|television| \\
    \hline
  \end{tabular}
  \caption{PASCAL-5\textsuperscript{i} tasks with FSS-1000 analog.}
  \label{table:pascal-5-tasks}
\end{center}
\end{table}

\section{FP-k Experimental Details}
\label{FP-k_Experimental_Details}
In this section, we describe our testing protocol evaluating the initializations when adapting to test tasks from the FP-k dataset. For each tuple of (initialization, k-training shots) we randomly sample 20 examples for a test set $\mathcal{D}^{test}$ for the task and train on k labeled examples $\mathcal{D}^{tr}$. We repeat this random sampling and training process 4 times for each of the 5 tasks, yielding 20 evaluation samples per (initialization, k-training shots) tuple. For all three initializations, we use UHO for estimating the hyperparameters of $U$ for $k < 10$. For $k \ge 10$, we use a fixed learning rate equal to the value used during meta-training and early stopping to estimate the optimal number of iterations. For early stopping, we use 20\% of the examples in $\mathcal{D}^{tr}$ to form $\mathcal{D}^{val}$.

\section{Proof of Generalization Gap}
\label{gen-gap}
The generalization gap is the difference between an estimate of the error of a function on an empirical dataset and the (typically non-computable) error over the true distribution \cite{kawaguchi2017generalization}. We define the generalization gap as the difference between the expected loss a model $f$ incurs over the true distribution $p$ and the loss measured on a dataset $\hat{p}$

\begin{equation}
  \E_p \left[\Ls \left( \hat{f} \right) \right] - \E_{\hat{p}} \left[\Ls \left( \hat{f} \right) \right]
\end{equation}

In meta-learning, $\hat{f}$ is learned on a distribution of examples $\hat{q}_{\tau}$ sampled from a distribution over tasks $\hat{p}$. Thus there is a function $\hat{f}_{\tau}$ that is learned on each $\hat{q}_{\tau}$

\begin{equation}
  \E_{p} \left[ \E_{q_{\tau}} \left[ \Ls \left( \hat{f}_{\tau}  \right) \right] \right] - \E_{\hat{p}} \left[ \E_{\hat{q}_{\tau}} \left[ \Ls \left( \hat{f}_{\tau}  \right) \right] \right]
\end{equation}

Without loss of generality, we can define an update operator $U$ which maps from a training distribution $q_{\tau}(x, y)$ and a parameter vector $\theta$ to a function $\hat{f}_{\tau}$:

\begin{equation}
\hat{f}_{\tau} = U(q_{\tau} ; \theta)
\end{equation}
To preserve generality, $U$ can be any arbitrary operator that returns a function $\hat{f}_{\tau}$. Replacing $\hat{f}_{\tau}$ with $U$ and dropping $q_{\tau}(x, y)$ for brevity:

\begin{equation}
	\E_{p} \left[ \E_{q_{\tau}} \left[ \Ls \left( U (\theta)  \right) \right] \right] - \E_{\hat{p}} \left[ \E_{\hat{q}_{\tau}} \left[ \Ls \left( U (\theta)  \right) \right] \right]
\end{equation}

We can, further, use a meta-learning algorithm to learn an initialization $\hat{\theta}^{*}$ that we estimate to be optimal  on some dataset of tasks $\mathcal{T}$

\begin{equation}
	\E_{p} \left[ \E_{q_{\tau}} \left[ \Ls \left( U (\hat{\theta}^{*})  \right) \right] \right] - \E_{\hat{p}} \left[ \E_{\hat{q}_{\tau}} \left[ \Ls \left( U (\hat{\theta}^{*})  \right) \right] \right].
\end{equation}

\section{Analysis of Weight Updates}
In this section we present empirical evidence that gradient-based mete-learning algorithms converge to a point in parameter space 
that is close in expectation to each task $\tau$'s manifold of optimal solutions for $\tau \in \mathcal{T}$. This builds on the theoretical analysis in Section 5.2 of \cite{nichol2018reptile}. First we compute the Euclidean distance between the entire EfficientLab-3 parameter vectors from an initialization $\theta$ to an updated weight vector $\theta_{\tau}$ after 5 gradient steps on 5 training examples $\mathcal{D}^{tr}$ from meta-test tasks $\tau \in \mathcal{T}^{test}$:

\begin{equation}
d_1 =  \left\| \theta - \theta_{\tau} \right\|_{2}
\end{equation}

We compute this distance twice on a random train-test split of the 10 examples for all 240 FSS-1000 test tasks, yielding 480 updated weight vector samples for each of the two meta-learned and joint-trained initializations.

\begin{figure}[H]
\begin{floatrow}
\centering
\capbtabbox{%
  \centering
  \begin{tabular}{cc} \hline
  Initialization method & $E_{\mathcal{T}^{test}}[d_1]$ \\ \hline
  Joint-trained & $0.995 \pm 0.022$ \\
  Meta-learned & $0.169 \pm 0.008$ \\ \hline
  \end{tabular}
}{}
\includegraphics[width=0.48\linewidth]{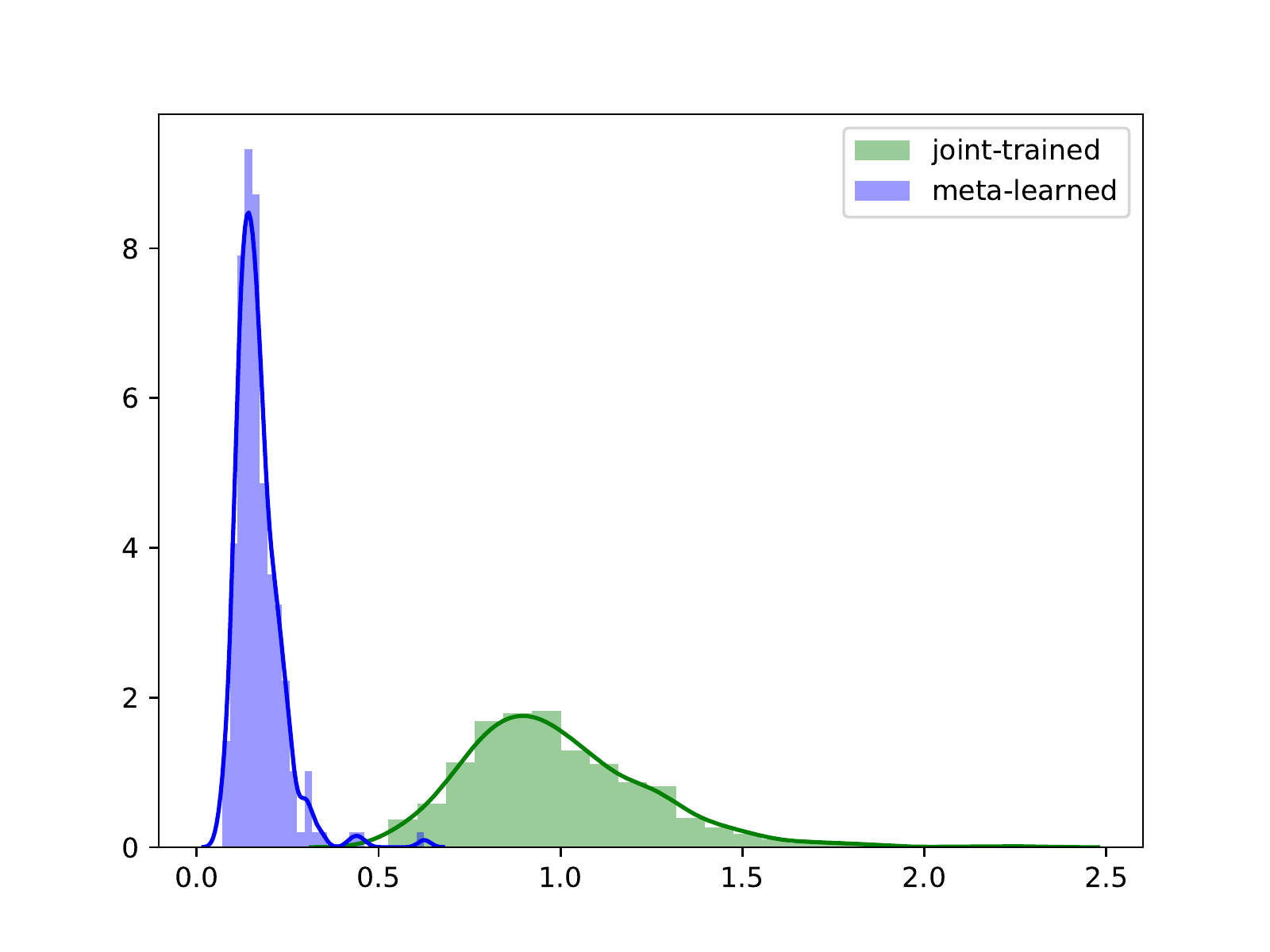}{}
\caption{{\bf Left:} Average Euclidean distance between initialization and updated weights with 95\% confidence interval. {\bf Right:} Distributions of Euclidean distances between initialization and updated weights.}
\end{floatrow}
\end{figure}

\begin{figure}[H]
  \centering
  \subfloat{\includegraphics[width=0.33\linewidth]{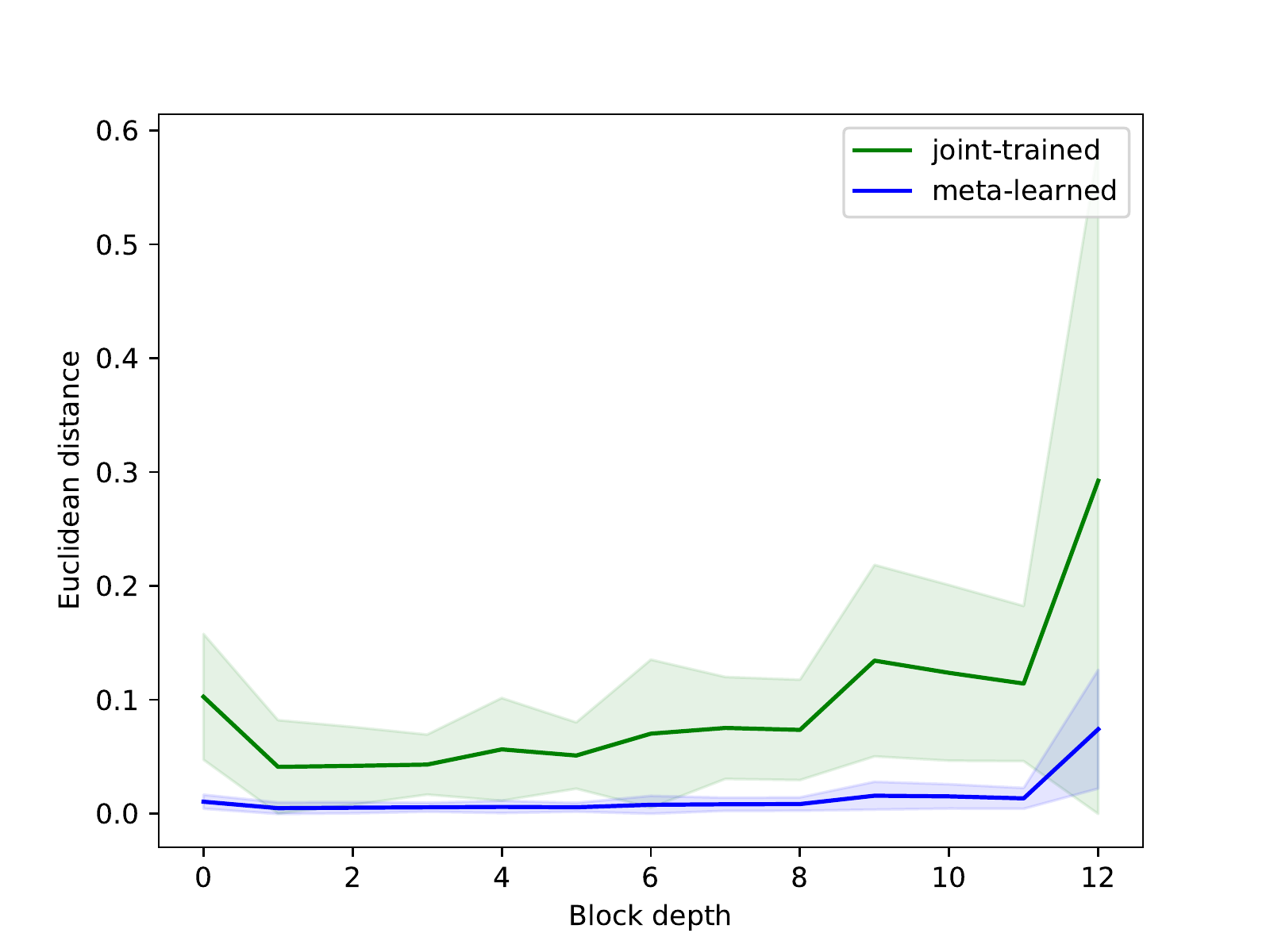}}
  \subfloat{\includegraphics[width=0.33\linewidth]{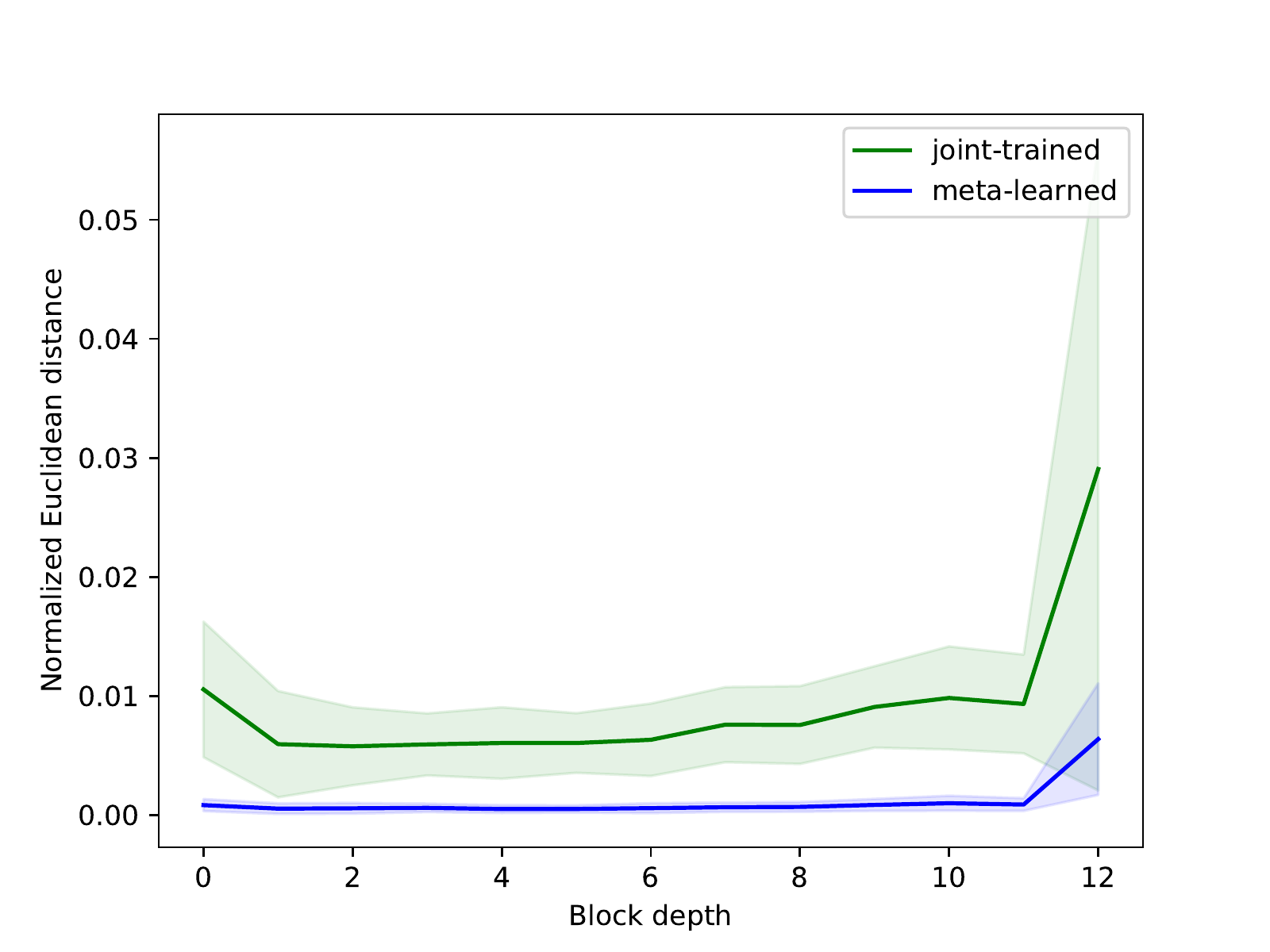}}
  \subfloat{\includegraphics[width=0.33\linewidth]{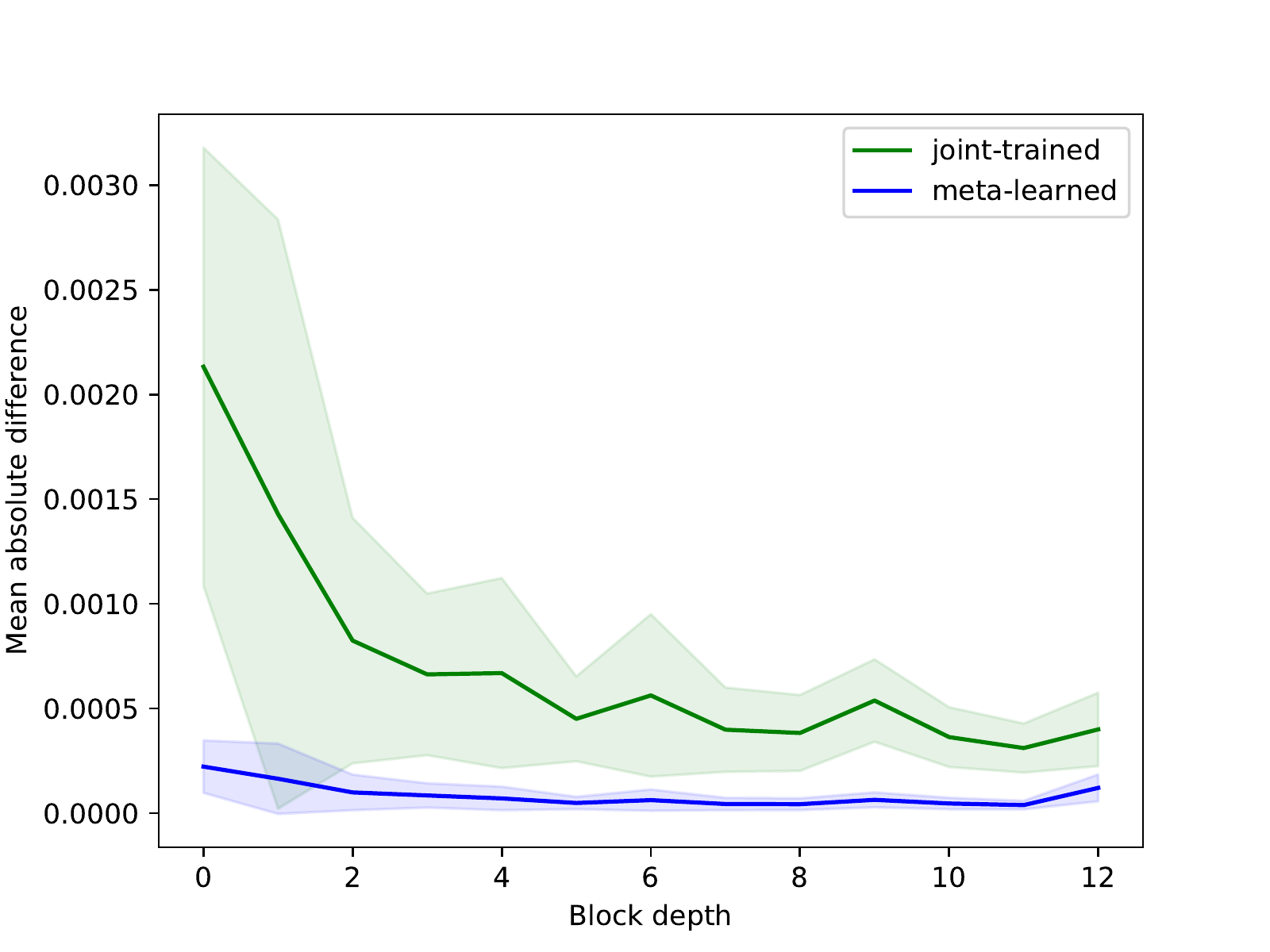}}
    \caption{Differences between weights at initialization and after 5 gradient steps with a step size of 0.005 on 5 training examples $\mathcal{D}^{tr}$ from test tasks $\mathcal{T}^{test}$. The left column shows the Euclidean distance ($d_1$) between EfficientLab-3 block weight vectors before and after training on $\mathcal{D}^{tr}$. Middle column shows Euclidean distance between unit norm weight vectors ($d_2$). Right column shows the mean absolute difference ($d_3$) between individual parameters in an EfficientLab block.}
    \label{fig:weights}
\end{figure}

Let $\mathbf{v} \in \mathbb{R}^{n}$ be a weight vector that is a subvector from an initialization $\theta$ and $\mathbf{u}_{\tau}  \in \mathbb{R}^{n}$ be the updated weight vector after gradient steps on examples sampled from $q_{\tau}$.  The subvectors $\mathbf{v}$ and $\mathbf{u}_{\tau}$ represent the weight tensors from an EfficientLab block unrolled into a vector. In Figure \ref{fig:weights}, we show the unit-normalized Euclidean distance between EfficientLab blocks, where a block is either the stem convolutional block, a mobile inverted bottleneck convolutional block~\cite{tan2019efficientnet,sandler2018mobilenetv2}, or our residual skip decoder:

\begin{equation}
d_2 = \left\| \frac{\mathbf{v}}{\left\| \mathbf{v} \right\|_{2}} - \frac{\mathbf{u}_{\tau}}{\left\| \mathbf{u}_{\tau} \right\|_{2}} \right\|_{2}
\end{equation}

We also plot the mean absolute difference to get a sense of the absolute distance traveled by individual parameters:
\begin{equation}
d_3 = \frac{1}{n}\sum_{i = 1}^{n} | \mathbf{v} - \mathbf{u}_{\tau} |_i
\end{equation}

As shown in Figure \ref{fig:weights}, we find that the joint-trained initialization travels significantly further when adapted to tasks from $\mathcal{T}^{test}$, even though the same learning rate and number of gradient steps are used at test time for both initializations. This implies that stable minima that produce low error lie closer in expectation over tasks from $\mathcal{T}^{test}$ to the meta-learned initialization. Evidence for this interpretation is further found in the test time metrics when evaluating with 5 gradient steps which show that the meta-learned initialization has a pixel-error rate that is 3.6 times smaller. This difference in error rate is found by comparing the joint-trained and FOMAML\textsuperscript{*} methods in Table 2.

The results in Figure \ref{fig:weights}, show that the adaptation to new tasks for both pre-training methods is non-uniform. The largest relative changes are shown in the final layers due to changes in the directionality of the EfficientLab weight subvectors. In contrast, the largest absolute changes in individual parameter values are found in the early layers of the EfficientLab model. Both initializations demonstrate similar patterns in the distribution of weight updates as a function of block depth but the changes in weights are up to an order of magnitude higher for the joint-trained initialization for all three difference metrics we investigated. The large difference between joint-trained and meta-learned distances is also in line with recent results of \cite{bengio2019meta} that show that when the knowledge of a model is factorized properly, the expected gradient over parameters when adapting to new tasks will be closer to zero.

\end{appendices}

\end{document}